# A Universal Variance Reduction-Based Catalyst for Nonconvex Low-Rank Matrix Recovery

Lingxiao Wang[*][‡]  and  Xiao Zhang[†][‡]  and  Quanquan Gu[§]

## Abstract

We propose a generic framework based on a new stochastic variance-reduced gradient descent algorithm for accelerating nonconvex low-rank matrix recovery. Starting from an appropriate initial estimator, our proposed algorithm performs projected gradient descent based on a novel semi-stochastic gradient specifically designed for low-rank matrix recovery. Based upon the mild restricted strong convexity and smoothness conditions, we derive a projected notion of the restricted Lipschitz continuous gradient property, and prove that our algorithm enjoys linear convergence rate to the unknown low-rank matrix with an improved computational complexity. Moreover, our algorithm can be employed to both noiseless and noisy observations, where the optimal sample complexity and the minimax optimal statistical rate can be attained respectively. We further illustrate the superiority of our generic framework through several specific examples, both theoretically and experimentally.

## 1 Introduction

Low-rank matrix recovery problem has been extensively studied during the past decades, due to its wide range of applications, such as collaborative filtering (Srebro et al., 2004; Rennie and Srebro, 2005) and multi-label learning (Cabral et al., 2011; Xu et al., 2013). The objective of low-rank matrix recovery is to recover the unknown low-rank matrix $\mathbf{X}^* \in \mathbb{R}^{d_1 \times d_2}$ from partial observations, such as a set of linear measurements in matrix sensing or a subset of its entries in matrix completion. Significant efforts have been made to estimate low-rank matrices, among which one of the most prevalent approaches is nuclear norm relaxation based optimization (Srebro et al., 2004; Candès and Tao, 2010; Rohde et al., 2011; Recht et al., 2010; Negahban and Wainwright, 2011, 2012; Gui and Gu, 2015). While such convex relaxation based methods enjoy a rigorous theoretical guarantee to recover the unknown low-rank matrix, due to the nuclear norm regularization/minimization, these algorithms involve a singular value decomposition at each iteration, whose time complexity is $O(d^3)$ to recover a $d \times d$ matrix. Therefore, they are computationally very expensive or even intractable.

In order to address the aforementioned computational issue, recent studies (Keshavan et al., 2010; Jain et al., 2013; Jain and Netrapalli, 2014; Zhao et al., 2015; Chen and Wainwright, 2015; Sun and Luo, 2015; Zheng and Lafferty, 2015, 2016; Tu et al., 2015; Bhojanapalli et al., 2015; Park et al., 2016b; Wang et al., 2016) have been carried out to perform factorization on the matrix space, which naturally ensures the low-rankness of the produced estimator. Although this matrix factorization technique converts the previous optimization problem into a nonconvex one, which is more difficult to deal with, it significantly improves the computational efficiency.

---
[*]Department of Systems and Information Engineering, University of Virginia, Charlottesville, VA 22904, USA; e-mail: lw4wr@virginia.edu

[†]Department of Statistics, University of Virginia, Charlottesville, VA 22904, USA; e-mail: xz7bc@virginia.edu

[‡]Equal Contribution

[§]Department of Systems and Information Engineering, Department of Computer Science, University of Virginia, Charlottesville, VA 22904, USA; e-mail: qg5w@virginia.edu

However, for large scale matrix recovery, such nonconvex optimization approaches are still computationally expensive, because they are based on gradient descent or alternating minimization, which involve the time-consuming calculation of full gradient at each iteration. De Sa et al. (2014) developed a stochastic gradient descent approach for Gaussian ensembles, but the sample complexity (i.e., number of measurements or observations required for exact recovery) of their algorithm is not optimal. Recently, Jin et al. (2016) and Zhang et al. (2017) proposed stochastic gradient descent algorithms for noiseless matrix completion and matrix sensing, respectively. Although these algorithms achieve linear rate of convergence and improved computational complexity over aforementioned deterministic optimization based approaches, they are limited to specific low-rank matrix recovery problems, and unable to be extended to more general problems and settings.

In this paper, in light of the idea of stochastic variance-reduced gradient (Johnson and Zhang, 2013), we propose a unified stochastic gradient descent framework with variance reduction for low-rank matrix recovery, which integrates both optimization-theoretic and statistical analyses. Inspired by the original meaning of catalyst in chemistry, we call our generic accelerating approach a 'catalyst'. To the best of our knowledge, this is the first unified accelerated stochastic gradient descent framework for low-rank matrix recovery with strong convergence guarantees. With a desired initial estimator given by a general initialization algorithm, we show that our algorithm achieves linear convergence rate and better computational complexity against the state-of-the-art algorithms.

## 1.1 Our Contributions

The contributions of our work are summarized as follows:

- We propose a unified accelerated stochastic gradient descent framework using variance reduction technique for low-rank matrix recovery. We develop a generic algorithm, which can be employed to both noisy and noiseless observations. In particular, for noisy observations, it is guaranteed to linearly converge to the unknown low-rank matrix up to the minimax statistical precision (Negahban and Wainwright, 2012; Koltchinskii et al., 2011; Wang et al., 2016). While in the noiseless case, our algorithm achieves the optimal sample complexity (Recht et al., 2010; Tu et al., 2015; Wang et al., 2016), while ensuring the linear rate of convergence. In addition, our general framework can be applied to a wide range of low rank-matrix estimation problems, including matrix sensing, matrix completion and one-bit matrix completion.

- At the core of our algorithm, we construct a novel semi-stochastic gradient term, which is substantially different from the one if following the original stochastic variance-reduced gradient using chain rule (Johnson and Zhang, 2013). This uniquely constructed semi-stochastic gradient has not been appeared in the literature, and is essential for deriving the minimax optimal statistical rate.

- Our unified framework is built upon the mild restricted strong convexity and smoothness conditions (Negahban et al., 2009; Negahban and Wainwright, 2011) regarding the objective function. Based on the above mentioned conditions, we derive an innovative projected notion of the restricted Lipschitz continuous gradient property, which we believe is of independent interest for other nonconvex problems to prove sharp statistical rates. We further establish the linear convergence rate of our generic algorithm. Besides, for each specific examples, we verify that the conditions required in our generic setting are satisfied with high probability to show the applicability of our framework.

- Our algorithm has a lower computational complexity compared with existing approaches (Jain et al., 2013; Zhao et al., 2015; Chen and Wainwright, 2015; Zheng and Lafferty, 2015, 2016; Tu et al., 2015; Bhojanapalli et al., 2015; Park et al., 2016b; Wang et al., 2016). More specifically, to achieve $\epsilon$ precision, the running time of our algorithm is $O\big((Nd'^2 + \kappa^2 bd'^2)\log(1/\epsilon)\big)$. Here $N$ denotes the total number of observations, $d'$ denotes the dimensionality of the unknown low-rank matrix $\mathbf{X}^*$, $b$ denotes the batch size, and $\kappa$ denotes the condition number of $\mathbf{X}^*$ (see Section 3 for a detailed definition). In particular, if the condition number satisfies $\kappa \leq N/b$, our algorithm is computationally more efficient than the state-of-the-art generic algorithm in Wang et al. (2016).

## 1.2 Organization and Notation

The remainder of this paper is organized as follows. In Section 2, we briefly review some related work. In Section 3, we present our stochastic variance-reduced gradient algorithm, and demonstrate its application to several specific matrix recovery examples. In Section 4, we first lay out necessary conditions for our framework, and then establish theoretical guarantees for our generic algorithm as well as specific examples. The main theoretical proofs are presented in Section 5. Numerical results and conclusion are provided in Sections 6 and 7 respectively. Appendix contains all the remaining proofs and technical lemmas.

Capital symbols such as $\mathbf{A}$ are used to denote matrices. We use $[d]$ and $\mathbf{I}_d$ to denote $\{1, 2, \ldots, d\}$ and $d \times d$ identity matrix respectively. We write $\mathbf{A}^\top \mathbf{A} = \mathbf{I}_{d_2}$, if matrix $\mathbf{A} \in \mathbb{R}^{d_1 \times d_2}$ is orthonormal. For any two matrices $\mathbf{A}$ and $\mathbf{B}$, we let $\langle \mathbf{A}, \mathbf{B} \rangle = \text{Tr}(\mathbf{A}^\top \mathbf{B})$ be the inner product between them. For any matrix $\mathbf{A} \in \mathbb{R}^{d_1 \times d_2}$, we use $\mathbf{A}_{i,*}$ and $\mathbf{A}_{*,j}$ to denote the $i$-th row and $j$-th column of $\mathbf{A}$, respectively. In addition, we use $A_{ij}$ to denote the $(i,j)$-th element of $\mathbf{A}$. Denote the row space and column space of $\mathbf{A}$ by row$(\mathbf{A})$ and col$(\mathbf{A})$ respectively. Let $d' = \max\{d_1, d_2\}$, and $\sigma_\ell(\mathbf{A})$ be the $\ell$-th largest singular value of $\mathbf{A}$. For any matrix $\mathbf{A} \in \mathbb{R}^{(d_1+d_2) \times r}$, denote the top $d_1 \times r$ and bottom $d_2 \times r$ matrices of $\mathbf{A}$ by $\mathbf{A}_U$ and $\mathbf{A}_V$ respectively. For a $d$-dimensional vector $\mathbf{x} = [x_1, x_2, \cdots, x_d]^\top \in \mathbb{R}^d$, we use $\|\mathbf{x}\|_q = (\Sigma_{i=1}^d |x_i|^q)^{1/q}$ to denote its $\ell_q$ vector norm for $0 < q < \infty$. As usual, we use $\|\mathbf{A}\|_2$ and $\|\mathbf{A}\|_F$ to denote the spectral norm and the Frobenius norm of matrix $\mathbf{A}$ respectively. Moreover, we use $\|\mathbf{A}\|_* = \sum_{i=1}^r \sigma_i(\mathbf{A})$ to denote the nuclear norm of $\mathbf{A}$, where $r$ is the rank of matrix $\mathbf{A}$, and use $\|\mathbf{A}\|_\infty = \max_{i,j} |A_{ij}|$ to denote the element-wise infinity norm of $\mathbf{A}$. Given two sequences $\{a_n\}$ and $\{b_n\}$, we write $a_n = O(b_n)$ if there exists a constant $0 < C_1 < \infty$ such that $a_n \leq C_1 b_n$.

## 2 Related Work

In this section, we briefly review some related work for nonconvex low-rank matrix recovery, and discuss some acceleration methods for stochastic optimization that are relevant to our work.

## 2.1 Nonconvex Low-Rank Matrix Recovery

Motivated by the empirical success of nonconvex optimization, a series of work was devoted to developing nonconvex optimization algorithms with theoretical guarantees for low-rank matrix recovery, such as alternating minimization (Jain et al., 2013; Hardt, 2014; Hardt and Wootters, 2014; Hardt et al., 2014; Zhao et al., 2015), gradient descent (Keshavan et al., 2009, 2010; Sun and Luo, 2015; Zheng and Lafferty, 2015, 2016; Tu et al., 2015; Bhojanapalli et al., 2015; Park et al., 2016a,b; Wang et al., 2016), and projected gradient descent (Jain and Netrapalli, 2014; Chen and Wainwright, 2015). For example, for matrix sensing and matrix completion, Jain et al. (2013) provided an analysis of the convergence rate of alternating minimization based on the restricted isometry and incoherence properties. Hardt (2014) analyzed alternating minimization for matrix completion from the perspective of power method. Later on, Zhao et al. (2015) established a more unified analysis for the recovery of low-rank matrix by showing that a broad class of nonconvex optimization algorithms, which includes both gradient-type and alternating minimization methods, can correctly recover the unknown low-rank matrix. However, the theoretical proofs of these algorithms mentioned above rely on the use of sample splitting technique, which results in unsatisfied sample complexity. Recently, Zheng and Lafferty (2015, 2016) proposed to use gradient descent based method to estimate the unknown low-rank matrix for matrix sensing and matrix completion. They proved that if the initial estimator is close enough to the unknown low-rank matrix, their approach is guaranteed to converge to the global optimum with a linear convergence rate. In addition, their method has better performances than alternating minimization in terms of both sample complexity and computational complexity. More recently, for matrix sensing, Tu et al. (2015) established an improved analysis of gradient descent based approach by assuming a refined restricted isometry property on linear measurements and using a more sophisticated initialization algorithm compared with Zheng and Lafferty (2015). In the mean time, Chen and Wainwright (2015) proposed a unified framework based on projected gradient descent scheme to recover the unknown low-rank matrix, specifically for the positive semidefinite matrix, and provided the theoretical guarantees of their algorithm for matrix sensing and matrix completion under the restricted isometry and incoherence properties.

However, both the restricted isometry property and incoherence condition are more stringent than the restricted strong convexity and smoothness conditions (Negahban et al., 2009; Negahban and Wainwright, 2011), which are adopted in this work. For instance, the restricted isometry property can be easily violated by simple random correlated sub-Gaussian designs (Negahban and Wainwright, 2011). Therefore, some recent work (Bhojanapalli et al., 2015; Park et al., 2016b; Wang et al., 2016) proposed to use the gradient descent based method for more general nonconvex low-rank matrix recovery problem through the restricted strong convexity and smoothness conditions of the objective function. In particular, Bhojanapalli et al. (2015) established a factorized gradient descent algorithm, which is based on the positive semidefinite matrix factorization, for nonconvex optimization. Built upon Bhojanapalli et al. (2015), under the same conditions, Park et al. (2016b) provided the theoretical guarantees of the factorized gradient descent approach and extended their results to the more general rectangular matrix factorization problem. But their analyses are limited to the optimization perspective, and do not consider the noisy setting. Most recently, Wang et al. (2016) proposed a unified framework for nonconvex low-rank matrix estimation, which covers both noisy and noiseless observations.

## 2.2 Stochastic Gradient Descent and Its Acceleration

One of the biggest issue of the aforementioned deterministic optimization methods is their computational complexity. Because these approaches need to evaluate the full gradient of the objective at each iteration, they are computationally prohibitive for large scale low-rank matrix recovery problems. One solution is to use stochastic gradient descent instead of gradient descent, since in stochastic gradient descent, only one or a mini-batch of stochastic gradients is calculated (Nemirovski et al., 2009; Lan, 2012) at each iteration. Thus the per-iteration time complexity can be reduced dramatically. Nevertheless, stochastic gradient descent often has a sublinear rate of convergence because random sampling can incur additional variance in estimating the gradient, even when the objective function is both strongly convex and smooth. Therefore, lots of efforts have been made in recent years to accelerate the stochastic gradient descent to linear rate of convergence through variance reduction. One way is to use the stochastic average gradient (Schmidt et al., 2013; Konečný and Richtárik, 2013; Defazio et al., 2014a,b; Mairal, 2014), another way is to use the stochastic variance-reduced gradient (a.k.a., semi-stochastic gradient) (Johnson and Zhang, 2013; Xiao and Zhang, 2014; Konečný et al., 2014). Both of these two variance reduction techniques can significantly improve the convergence of stochastic gradient descent. In addition, unlike stochastic average gradient, stochastic variance-reduced gradient does not require the storage of stochastic gradients, which is more practical for large scale problems due to its low memory cost. Built upon stochastic variance-reduced gradient, Reddi et al. (2016); Allen-Zhu and Hazan (2016) established stochastic variance-reduced gradient descent algorithms for the general nonconvex finite-sum optimization problem, which are guaranteed to converge to the stationary point with a sublinear convergence rate. However, their nonconvex optimization algorithms and analyses can not be directly applied to low-rank matrix recovery and many other nonconvex optimization problems.

Therefore, many stochastic variance-reduced gradient descent algorithms have been invented for specific nonconvex optimization problems. For instance, Shamir (2015a,b) proposed and analyzed stochastic variance reduced power method for principal component analysis. Garber and Hazan (2015); Garber et al. (2016) proposed to first use shift-and-inverse preconditioning technique to reduce the problem to solve a sequence of linear systems, then they applied the stochastic variance-reduced gradient to solve the linear systems more efficiently. For sparsity constrained nonconvex statistical learning, Li et al. (2016) proposed a stochastic variance-reduced gradient hard thresholding algorithm to efficiently solve the high dimensional sparse learning problem under a cardinality constraint. Later, Chen and Gu (2016) invented a stochastic variance-reduced mini-batch randomized block coordinate gradient descent algorithm to further accelerate the convergence by leveraging the double randomization: mini-batch sampling and coordinate block sampling. For nonconvex low-rank matrix recovery, we note that De Sa et al. (2014) proposed a stochastic gradient descent algorithm, which can be seen as a stochastic power iteration, for matrix completion. Although they provided a global convergence result with a random initialization, the sample complexity of their method is very high. There is little work on accelerating nonconvex low-rank matrix recovery. The only exceptions we are aware of are Jin et al. (2016) and Zhang et al. (2017). In detail, Jin et al. (2016) proposed a provable, efficient online algorithm for matrix completion based upon nonconvex stochastic gradient descent. Although it can address the online matrix completion problem, their analysis is limited to matrix completion with noiseless observations. In addition, their algorithm has a higher sample complexity and sub-optimal convergence rate. Recently, Zhang et al. (2017) proposed a stochastic variance-reduced gradient descent algorithm for matrix sensing. Nevertheless, their analyses are based on the restricted isometry property, which is stronger than the conditions employed in this paper, and limited to matrix sensing.

## 3 Methodology

In this section, we first present the stochastic variance-reduced gradient descent algorithm for low-rank matrix recovery. Then we illustrate the applicability of our algorithm with several specific matrix recovery examples.

### 3.1 Stochastic Variance-Reduced Gradient for Low-Rank Matrix Recovery

Before we present our proposed stochastic variance-reduced gradient descent algorithm, we first briefly introduce the general problem setup for low-rank matrix recovery. Suppose $\mathbf{X}^* \in \mathbb{R}^{d_1 \times d_2}$ is an unknown rank-$r$ matrix. Let the singular value decomposition (SVD) of $\mathbf{X}^*$ be $\mathbf{X}^* = \overline{\mathbf{U}}^* \mathbf{\Sigma}^* \overline{\mathbf{V}}^{*\top}$, where $\overline{\mathbf{U}}^* \in \mathbb{R}^{d_1 \times r}$, $\overline{\mathbf{V}}^* \in \mathbb{R}^{d_2 \times r}$ are orthonormal matrices, and $\mathbf{\Sigma}^* \in \mathbb{R}^{r \times r}$ is a diagonal matrix. Let $\sigma_1 \geq \sigma_2 \geq \cdots \geq \sigma_r \geq 0$ be the sorted nonzero singular values of $\mathbf{X}^*$, and denote the condition number of $\mathbf{X}^*$ by $\kappa$, i.e., $\kappa = \sigma_1/\sigma_r$. Besides, let $\mathbf{U}^* = \overline{\mathbf{U}}^* (\mathbf{\Sigma}^*)^{1/2}$ and $\mathbf{V}^* = \overline{\mathbf{V}}^* (\mathbf{\Sigma}^*)^{1/2}$. Recall that we aim to recover $\mathbf{X}^*$ through a collection of $N$ observations or measurements. Let $\mathcal{L}_N : \mathbb{R}^{d_1 \times d_2} \to \mathbb{R}$ be the sample loss function, which evaluates the fitness of any matrix $\mathbf{X}$ associated with the total N observations. Then the low-rank matrix recovery problem can be formulated as follows:

$$\min_{\mathbf{X} \in \mathbb{R}^{d_1 \times d_2}} \mathcal{L}_N(\mathbf{X}) := \frac{1}{N} \sum_{i=1}^{N} \ell_i(\mathbf{X}), \text{ subject to } \mathbf{X} \in \mathcal{C}, \text{rank}(\mathbf{X}) \leq r, \tag{3.1}$$

where $\ell_i(\mathbf{X})$ measures the fitness of $\mathbf{X}$ associated with the $i$-th observation. Here $\mathcal{C} \subseteq \mathbb{R}^{d_1 \times d_2}$ is a feasible set, such that $\mathbf{X}^* \in \mathcal{C}$. Note that if $\mathcal{C} = \mathbb{R}^{d_1 \times d_2}$, optimization problem (3.1) degenerates to the standard one. In order to more efficiently estimate the unknown low-rank matrix, following Jain et al. (2013); Tu et al. (2015); Zheng and Lafferty (2016); Park et al. (2016a); Wang et al. (2016), we decompose $\mathbf{X}$ as $\mathbf{UV}^\top$ and consider the following nonconvex optimization problem via matrix factorization:

$$\min_{\substack{\mathbf{U} \in \mathbb{R}^{d_1 \times r} \\ \mathbf{V} \in \mathbb{R}^{d_2 \times r}}} \mathcal{L}_N(\mathbf{UV}^\top) := \frac{1}{N} \sum_{i=1}^{N} \ell_i(\mathbf{UV}^\top), \text{ subject to } \mathbf{U} \in \mathcal{C}_1, \mathbf{V} \in \mathcal{C}_2, \tag{3.2}$$

where $\mathcal{C}_1 \subseteq \mathbb{R}^{d_1 \times r}, \mathcal{C}_2 \subseteq \mathbb{R}^{d_2 \times r}$ are the rotation-invariant sets induced by $\mathcal{C}$. Suppose $\mathbf{X}^*$ can be factorized as $\mathbf{X}^* = \mathbf{U}^* \mathbf{V}^{*\top}$, then we need to make sure that $\mathbf{U}^* \in \mathcal{C}_1$ and $\mathbf{V}^* \in \mathcal{C}_2$. Besides, it can be seen from (3.2) that the optimal solution is not unique in terms of rotation. In order to deal with such identifiability issue, following Tu et al. (2015); Zheng and Lafferty (2016); Park et al. (2016b), we consider the following regularized optimization problem:

$$\min_{\substack{\mathbf{U} \in \mathbb{R}^{d_1 \times r} \\ \mathbf{V} \in \mathbb{R}^{d_2 \times r}}} F_N(\mathbf{U}, \mathbf{V}) := \mathcal{L}_N(\mathbf{UV}^\top) + \frac{1}{8} \|\mathbf{U}^\top \mathbf{U} - \mathbf{V}^\top \mathbf{V}\|_F^2, \quad \text{subject to} \quad \mathbf{U} \in \mathcal{C}_1, \mathbf{V} \in \mathcal{C}_2. \tag{3.3}$$

Besides, we rewrite the objective function $F_N(\mathbf{U}, \mathbf{V})$ as a sum of $n$ components to make use of the idea of stochastic gradient descent with variance reduction:

$$F_N(\mathbf{U}, \mathbf{V}) := \frac{1}{n} \sum_{i=1}^{n} F_i(\mathbf{U}, \mathbf{V}), \tag{3.4}$$

where we assume $N = nb$, and $b$ denotes batch size, i.e., the number of observations associated with each $F_i$. More specifically, we have

$$F_i(\mathbf{U}, \mathbf{V}) = \mathcal{L}_i(\mathbf{U}\mathbf{V}^\top) + \frac{1}{8}\|\mathbf{U}^\top\mathbf{U} - \mathbf{V}^\top\mathbf{V}\|_F^2, \quad \text{where } \mathcal{L}_i(\mathbf{U}\mathbf{V}^\top) = \frac{1}{b}\sum_{j=1}^{b} \ell_{i_j}(\mathbf{U}\mathbf{V}^\top). \quad (3.5)$$

Therefore, based on (3.4) and (3.5), we are able to apply the stochastic variance-reduced gradient, which is displayed as Algorithm 1. As will be seen in later theoretical analysis, the variance of the proposed stochastic gradient indeed decreases as the iteration number increases, which leads to faster convergence rate. We let $\mathcal{P}_{\mathcal{C}_i}$ be the projection operator onto the feasible set $\mathcal{C}_i$ in Algorithm 1, where $i \in \{1, 2\}$.

---

**Algorithm 1** Stochastic Variance-Reduced Gradient for Low-Rank Matrix Recovery

**Input:** loss function $\mathcal{L}_N$; step size $\eta$; number of iterations $S, m$; initial solution $(\widetilde{\mathbf{U}}^0, \widetilde{\mathbf{V}}^0)$.
    **for:** $s = 1, 2, \ldots, S$ **do**
        $\widetilde{\mathbf{U}} = \widetilde{\mathbf{U}}^{s-1}, \widetilde{\mathbf{V}} = \widetilde{\mathbf{V}}^{s-1}, \widetilde{\mathbf{X}} = \widetilde{\mathbf{U}}\widetilde{\mathbf{V}}^\top$
        $\mathbf{U}^0 = \widetilde{\mathbf{U}}, \mathbf{V}^0 = \widetilde{\mathbf{V}}$
        **for:** $t = 0, 1, 2, \ldots, m-1$ **do**
            Randomly pick $i_t \in \{1, 2, \ldots, n\}$
            $\mathbf{U}^{t+1} = \mathbf{U}^t - \eta\bigl(\nabla_\mathbf{U} F_{i_t}(\mathbf{U}^t, \mathbf{V}^t) - \nabla\mathcal{L}_{i_t}(\widetilde{\mathbf{X}})\mathbf{V}^t + \nabla\mathcal{L}_N(\widetilde{\mathbf{X}})\mathbf{V}^t\bigr)$
            $\mathbf{V}^{t+1} = \mathbf{V}^t - \eta\bigl(\nabla_\mathbf{V} F_{i_t}(\mathbf{U}^t, \mathbf{V}^t) - \nabla\mathcal{L}_{i_t}(\widetilde{\mathbf{X}})^\top\mathbf{U}^t + \nabla\mathcal{L}_N(\mathbf{X}^t)^\top\mathbf{U}^t\bigr)$
            $\mathbf{U}^{t+1} = \mathcal{P}_{\mathcal{C}_1}(\mathbf{U}^{t+1})$
            $\mathbf{V}^{t+1} = \mathcal{P}_{\mathcal{C}_2}(\mathbf{V}^{t+1})$
        **end for**
        $\widetilde{\mathbf{U}}^s = \mathbf{U}^t, \widetilde{\mathbf{V}}^s = \mathbf{V}^t$ for randomly chosen $t \in \{0, \ldots, m-1\}$
    **end for**
**Output:** $(\widetilde{\mathbf{U}}^S, \widetilde{\mathbf{V}}^S)$.

---

Note that our proposed Algorithm 1 is different from the standard stochastic variance-reduced gradient algorithm (Johnson and Zhang, 2013) in several aspects. First, we introduce a projection step to ensure that the estimator produced at each iteration belongs to a feasible set, which is necessary for various low-rank matrix recovery problems. Second, we construct a novel semi-stochastic gradient term for $\mathbf{U}$ (resp. $\mathbf{V}$) as $\nabla_\mathbf{U} F_{i_t}(\mathbf{U}, \mathbf{V}) - \nabla\mathcal{L}_{i_t}(\widetilde{\mathbf{X}})\mathbf{V} + \nabla\mathcal{L}_N(\widetilde{\mathbf{X}})\mathbf{V}$, instead of $\nabla_\mathbf{U} F_{i_t}(\mathbf{U}, \mathbf{V}) - \nabla_\mathbf{U} F_{i_t}(\widetilde{\mathbf{U}}, \widetilde{\mathbf{V}}) + \nabla_\mathbf{U} F_N(\widetilde{\mathbf{U}}, \widetilde{\mathbf{V}})$ if following the original stochastic variance reduced gradient descent (Johnson and Zhang, 2013). This uniquely devised semi-stochastic gradient is essential for deriving the minimax optimal statistical rate.

Besides, following Wang et al. (2016), we propose an initialization algorithm, which is shown in Algorithm 2, to ensure the initial solution $(\widetilde{\mathbf{U}}^0, \widetilde{\mathbf{V}}^0)$ is close enough to $(\mathbf{U}^*, \mathbf{V}^*)$, which is essential to derive the linear convergence rate of Algorithm 1. For any rank-$r$ matrix $\mathbf{X} \in \mathbb{R}^{d_1 \times d_2}$, we use $\text{SVD}_r(\mathbf{X})$ to denote its singular value decomposition. If $\text{SVD}_r(\mathbf{X}) = [\mathbf{U}, \mathbf{\Sigma}, \mathbf{V}]$, then we use $\mathcal{P}_r(\mathbf{X}) = \mathbf{U}\mathbf{\Sigma}\mathbf{V}^\top$ to denote the best rank-$r$ approximation of $\mathbf{X}$, or in other words, $\mathcal{P}_r : \mathbb{R}^{d_1 \times d_2} \to \mathbb{R}^{d_1 \times d_2}$ denotes the projection operator such that $\mathcal{P}_r(\mathbf{X}) = \text{argmin}_{\text{rank}(\mathbf{Y}) \leq r} \|\mathbf{X} - \mathbf{Y}\|_F$.

**Algorithm 2** Initialization
> **Input:** loss function $\mathcal{L}_N$; step size $\tau$; number of iterations T.
> $\mathbf{X}_0 = \mathbf{0}$
> **for:** $t = 1, 2, 3, \ldots, T$ **do**
>   $\mathbf{X}_t = \mathcal{P}_r(\mathbf{X}_{t-1} - \tau \nabla \mathcal{L}_N(\mathbf{X}_{t-1}))$
> **end for**
> $[\overline{\mathbf{U}}^0, \mathbf{\Sigma}^0, \overline{\mathbf{V}}^0] = \text{SVD}_r(\mathbf{X}_T)$
> $\widetilde{\mathbf{U}}^0 = \overline{\mathbf{U}}^0 (\mathbf{\Sigma}^0)^{1/2}, \widetilde{\mathbf{V}}^0 = \overline{\mathbf{V}}^0 (\mathbf{\Sigma}^0)^{1/2}$
> **Output:** $(\widetilde{\mathbf{U}}^0, \widetilde{\mathbf{V}}^0)$

Combining stochastic variance-reduced gradient Algorithm 1 and initialization Algorithm 2, it is guaranteed that the estimator $(\widetilde{\mathbf{U}}^S, \widetilde{\mathbf{V}}^S)$ produced in Algorithm 1 will converge to $(\mathbf{U}^*, \mathbf{V}^*)$ in expectation.

### 3.2 Applications to Specific Models

In this subsection, we introduce three examples, which include matrix sensing, matrix completion and one-bit matrix completion, to illustrate the applicability of our proposed algorithm (Algorithm 1). To apply the proposed method, we only need to specify the form of $F_N(\mathbf{U}, \mathbf{V})$ for each specific model, as defined in (3.4).

#### 3.2.1 Matrix Sensing

In matrix sensing (Recht et al., 2010; Negahban and Wainwright, 2011), we intend to recover the unknown matrix $\mathbf{X}^* \in \mathbb{R}^{d_1 \times d_2}$ with rank-$r$ from a set of noisy linear measurements such that $\mathbf{y} = \mathcal{A}(\mathbf{X}^*) + \boldsymbol{\epsilon}$, where the linear measurement operator $\mathcal{A} : \mathbb{R}^{d_1 \times d_2} \to \mathbb{R}^N$ is defined as $\mathcal{A}(\mathbf{X}) = (\langle \mathbf{A}_1, \mathbf{X} \rangle, \langle \mathbf{A}_2, \mathbf{X} \rangle, \ldots, \langle \mathbf{A}_N, \mathbf{X} \rangle)^\top$, for any $\mathbf{X} \in \mathbb{R}^{d_1 \times d_2}$. Here $N$ denotes the number of observations, and $\boldsymbol{\epsilon}$ represents a sub-Gaussian noise vector with i.i.d. elements and parameter $\nu$. In addition, for each sensing matrix $\mathbf{A}_i \in \mathbb{R}^{d_1 \times d_2}$, it has i.i.d. standard Gaussian entries. Therefore, we formulate $F_N(\mathbf{U}, \mathbf{V})$ for matrix sensing as follows

$$F_N(\mathbf{U}, \mathbf{V}) = \frac{1}{2N} \|\mathbf{y} - \mathcal{A}(\mathbf{U}\mathbf{V}^\top)\|_2^2 + \frac{1}{8} \|\mathbf{U}^\top \mathbf{U} - \mathbf{V}^\top \mathbf{V}\|_F^2 = \frac{1}{n} \sum_{i=1}^n F_{\mathcal{S}_i}(\mathbf{U}, \mathbf{V}).$$

For each component function, we have

$$F_{\mathcal{S}_i}(\mathbf{U}, \mathbf{V}) = \frac{1}{2b} \|\mathbf{y}_{\mathcal{S}_i} - \mathcal{A}_{\mathcal{S}_i}(\mathbf{U}\mathbf{V}^\top)\|_2^2 + \frac{1}{8} \|\mathbf{U}^\top \mathbf{U} - \mathbf{V}^\top \mathbf{V}\|_F^2,$$

where $\{\mathcal{S}_i\}_{i=1}^n$ denote the mutually disjoint subsets such that $\cup_{i=1}^n \mathcal{S}_i = [N]$, and $\mathcal{A}_{\mathcal{S}_i}$ is defined as a linear measurement operator $\mathcal{A}_{\mathcal{S}_i} : \mathbb{R}^{d_1 \times d_2} \to \mathbb{R}^b$, satisfying $\mathcal{A}_{\mathcal{S}_i}(\mathbf{X}) = (\langle \mathbf{A}_{i_1}, \mathbf{X} \rangle, \langle \mathbf{A}_{i_2}, \mathbf{X} \rangle, \ldots, \langle \mathbf{A}_{i_b}, \mathbf{X} \rangle)^\top$, with corresponding observations $\mathbf{y}_{\mathcal{S}_i} = (y_{i_1}, y_{i_2}, \ldots, y_{i_b})^\top$.

#### 3.2.2 Matrix Completion

For matrix completion with noisy observations (Rohde et al., 2011; Koltchinskii et al., 2011; Negahban and Wainwright, 2012), our primary goal is to recover the unknown low-rank matrix $\mathbf{X}^* \in \mathbb{R}^{d_1 \times d_2}$ from a set of randomly observed noisy elements, which is generated from $\mathbf{X}^*$. For example, one commonly-used model is the uniform observation model, which is defined as follows

$$Y_{jk} := \begin{cases} X_{jk}^* + Z_{jk}, & \text{with probability } p, \\ *, & \text{otherwise,} \end{cases}$$

where $\mathbf{Z} \in \mathbb{R}^{d_1 \times d_2}$ is a noise matrix such that each element $Z_{jk}$ follows i.i.d. sub-Gaussian distribution with parameter $\nu$, and we call $\mathbf{Y} \in \mathbb{R}^{d_1 \times d_2}$ as the observation matrix. In particular, we observe each elements independently with probability $p \in (0,1)$. We denote $\Omega \subseteq [d_1] \times [d_2]$ by the index set of the observed entries, then $F_\Omega(\mathbf{U}, \mathbf{V})$ for matrix completion is formulated as follows

$$F_\Omega(\mathbf{U}, \mathbf{V}) := \frac{1}{2p} \sum_{(j,k) \in \Omega} (\mathbf{U}_{j*}\mathbf{V}_{k*}^\top - Y_{jk})^2 + \frac{1}{8}\|\mathbf{U}^\top\mathbf{U} - \mathbf{V}^\top\mathbf{V}\|_F^2 = \frac{1}{n}\sum_{i=1}^n F_{\Omega_{\mathcal{S}_i}}(\mathbf{U}, \mathbf{V}),$$

where $p = |\Omega|/(d_1 d_2)$, and each component function is defined as

$$F_{\Omega_{\mathcal{S}_i}}(\mathbf{U}, \mathbf{V}) = \frac{1}{2p'} \sum_{(j,k) \in \Omega_{\mathcal{S}_i}} (\mathbf{U}_{j*}\mathbf{V}_{k*}^\top - Y_{jk})^2 + \frac{1}{8}\|\mathbf{U}^\top\mathbf{U} - \mathbf{V}^\top\mathbf{V}\|_F^2.$$

Here $\{\Omega_{\mathcal{S}_i}\}_{i=1}^n$ denote the mutually disjoint subsets such that $\cup_{i=1}^n \Omega_{\mathcal{S}_i} = \Omega$. In addition, we have $|\Omega_{\mathcal{S}_i}| = b$ for $i = 1, \ldots, n$ such that $|\Omega| = nb$, and $p' = b/(d_1 d_2)$.

### 3.2.3 One-bit Matrix Completion

Compared with matrix completion, we only observe the sign of each noisy entries of the unknown low-rank matrix $\mathbf{X}^*$ in one-bit matrix completion (Davenport et al., 2014; Cai and Zhou, 2013). We consider the uniform sampling model, which has been studied in existing literature (Davenport et al., 2014; Cai and Zhou, 2013; Ni and Gu, 2016). More specifically, we consider the following observation model, which is based on a probability density function $f : \mathbb{R} \to [0,1]$

$$Y_{jk} = \begin{cases} +1, & \text{with probability } f(X_{jk}^*), \\ -1, & \text{with probability } 1 - f(X_{jk}^*), \end{cases} \tag{3.6}$$

where we use a binary matrix $\mathbf{Y}$ to denote the observation matrix in (3.6). In addition, if the probability density function $f$ is a cumulative distribution function with respect to $-Z_{jk}$, then we can rewrite the observation model (3.6) as follows

$$Y_{jk} = \begin{cases} +1, & \text{if } X_{jk}^* + Z_{jk} > 0, \\ -1, & \text{if } X_{jk}^* + Z_{jk} < 0, \end{cases} \tag{3.7}$$

where we use $\mathbf{Z} \in \mathbb{R}^{d_1 \times d_2}$ to denote the noise matrix with i.i.d. elements $Z_{jk}$. Lots of probability density functions can be applied to observation model (3.6), and we consider the broadly-used logistic function $f(X_{jk}) = e^{X_{jk}}/(1 + e^{X_{jk}})$ as the density function in our study, which is equivalent to the fact that each noise element $Z_{jk}$ in model (3.7) follows the standard logistic distribution. Similar to matrix completion, we use $\Omega \subseteq [d_1] \times [d_2]$ to denote the index set of the observed elements.

Therefore, given the logistic function $f$ and the index set $\Omega$, we define $F_\Omega(\mathbf{U}, \mathbf{V})$ for one-bit matrix completion as follows

$$F_\Omega(\mathbf{U}, \mathbf{V}) := \mathcal{L}_\Omega(\mathbf{U}\mathbf{V}^\top) + \frac{1}{8}\|\mathbf{U}^\top\mathbf{U} - \mathbf{V}^\top\mathbf{V}\|_F^2 = \frac{1}{n}\sum_{i=1}^n F_{\Omega_{\mathcal{S}_i}}(\mathbf{U}, \mathbf{V}),$$

where $\mathcal{L}_\Omega(\mathbf{U}\mathbf{V}^\top)$ is the negative log-likelihood function such that

$$\mathcal{L}_\Omega(\mathbf{U}\mathbf{V}^\top) = -\frac{1}{p}\sum_{(j,k)\in\Omega}\left\{\mathbb{1}_{(Y_{jk}=1)}\log\left(f(\mathbf{U}_{j*}\mathbf{V}_{*k}^\top)\right) + \mathbb{1}_{(Y_{jk}=-1)}\log\left(1 - f(\mathbf{U}_{j*}\mathbf{V}_{*k}^\top)\right)\right\},$$

where $p = |\Omega|/(d_1 d_2)$. Therefore, for each component function, we have

$$F_{\Omega_{\mathcal{S}_i}}(\mathbf{U}, \mathbf{V}) = \mathcal{L}_{\Omega_{\mathcal{S}_i}}(\mathbf{U}\mathbf{V}^\top) + \frac{1}{8}\|\mathbf{U}^\top\mathbf{U} - \mathbf{V}^\top\mathbf{V}\|_F^2,$$

where $\{\Omega_{\mathcal{S}_i}\}_{i=1}^n$ denote the mutually disjoint subsets such that $\cup_{i=1}^n \Omega_{\mathcal{S}_i} = \Omega$. In addition, we have $|\Omega_{\mathcal{S}_i}| = b$ for $i = 1, \ldots, n$ such that $|\Omega| = nb$. And $\mathcal{L}_{\Omega_{\mathcal{S}_i}}(\mathbf{U}\mathbf{V}^\top)$ is defined as

$$\mathcal{L}_{\Omega_{\mathcal{S}_i}}(\mathbf{U}\mathbf{V}^\top) = \frac{1}{p'}\sum_{(j,k)\in\Omega_{\mathcal{S}_i}}\left\{\mathbb{1}_{(Y_{jk}=1)}\log\left(f(\mathbf{U}_{j*}\mathbf{V}_{*k}^\top)\right) + \mathbb{1}_{(Y_{jk}=-1)}\log\left(1 - f(\mathbf{U}_{j*}\mathbf{V}_{*k}^\top)\right)\right\},$$

where $p' = b/(d_1 d_2)$.

## 4 Main Theory

In this section, we present our main theoretical results for Algorithms 1 and 2. We first introduce several definitions for simplicity.

Recall that the singular value decomposition of $\mathbf{X}^*$ is $\mathbf{X}^* = \overline{\mathbf{U}}^*\mathbf{\Sigma}^*\overline{\mathbf{V}}^{*\top}$, then following the (Tu et al., 2015; Zheng and Lafferty, 2016), we define $\mathbf{Y}^* \in \mathbb{R}^{(d_1+d_2)\times(d_1+d_2)}$ as the corresponding lifted positive semidefinite matrix of $\mathbf{X}^* \in \mathbb{R}^{d_1 \times d_2}$ in higher dimension

$$\mathbf{Y}^* = \begin{bmatrix} \mathbf{U}^*\mathbf{U}^{*\top} & \mathbf{U}^*\mathbf{V}^{*\top} \\ \mathbf{V}^*\mathbf{U}^{*\top} & \mathbf{V}^*\mathbf{V}^{*\top} \end{bmatrix} = \mathbf{Z}^*\mathbf{Z}^{*\top},$$

where $\mathbf{U}^* = \overline{\mathbf{U}}^*(\mathbf{\Sigma}^*)^{1/2}$, $\mathbf{V}^* = \overline{\mathbf{V}}^*(\mathbf{\Sigma}^*)^{1/2}$, and $\mathbf{Z}^*$ is defined as

$$\mathbf{Z}^* = \begin{bmatrix} \mathbf{U}^* \\ \mathbf{V}^* \end{bmatrix} \in \mathbb{R}^{(d_1+d_2)\times r}.$$

Besides, we define the solution set in terms of the true parameter $\mathbf{Z}^*$ as follows

$$\mathcal{Z} = \left\{\mathbf{Z} \in \mathbb{R}^{(d_1+d_2)\times r} \mid \mathbf{Z} = \mathbf{Z}^*\mathbf{R} \text{ for some } \mathbf{R} \in \mathbb{Q}_r\right\},$$

where $\mathbb{Q}_r$ denotes the set of $r \times r$ orthonormal matrices. According to this definition, for any $\mathbf{Z} \in \mathcal{Z}$, we can obtain $\mathbf{X}^* = \mathbf{Z}_U\mathbf{Z}_V^\top$, where $\mathbf{Z}_U$ and $\mathbf{Z}_V$ denote the top $d_1 \times r$ and bottom $d_2 \times r$ matrices of $\mathbf{Z} \in \mathbb{R}^{(d_1+d_2)\times r}$ respectively.

**Definition 4.1.** Define the distance between $\mathbf{Z}$ and $\mathbf{Z}^*$ in terms of the optimal rotation as $d(\mathbf{Z}, \mathbf{Z}^*)$ such that

$$d(\mathbf{Z}, \mathbf{Z}^*) = \min_{\widetilde{\mathbf{Z}} \in \mathcal{Z}} \|\mathbf{Z} - \widetilde{\mathbf{Z}}\|_F = \min_{\mathbf{R} \in \mathbb{Q}_r} \|\mathbf{Z} - \mathbf{Z}^* \mathbf{R}\|_F.$$

**Definition 4.2.** Define the neighbourhood of $\mathbf{Z}^*$ with radius $R$ as

$$\mathbb{B}(R) = \left\{ \mathbf{Z} \in \mathbb{R}^{(d_1+d_2) \times r} \,\middle|\, d(\mathbf{Z}, \mathbf{Z}^*) \leq R \right\}.$$

Next, we lay out several conditions, which is essential for proving our main theory. We impose restricted strong convexity (RSC) and smoothness (RSS) conditions (Negahban et al., 2009; Loh and Wainwright, 2013) on the sample loss function $\mathcal{L}_N$.

**Condition 4.3** (Restricted Strong Convexity). Given a fixed sample size $N$, assume $\mathcal{L}_N$ is restricted strongly convex with parameter $\mu$, such that for all matrices $\mathbf{X}, \mathbf{Y} \in \mathbb{R}^{d_1 \times d_2}$ with rank at most $3r$

$$\mathcal{L}_N(\mathbf{Y}) \geq \mathcal{L}_N(\mathbf{X}) + \langle \nabla \mathcal{L}_N(\mathbf{X}), \mathbf{Y} - \mathbf{X} \rangle + \frac{\mu}{2} \|\mathbf{Y} - \mathbf{X}\|_F^2.$$

**Condition 4.4** (Restricted Strong Smoothness). Given a fixed sample size $N$, assume $\mathcal{L}_N$ is restricted strongly smooth with parameter $L$, such that for all matrices $\mathbf{X}, \mathbf{Y} \in \mathbb{R}^{d_1 \times d_2}$ with rank at most $3r$

$$\mathcal{L}_N(\mathbf{Y}) \leq \mathcal{L}_N(\mathbf{X}) + \langle \nabla \mathcal{L}_N(\mathbf{X}), \mathbf{Y} - \mathbf{X} \rangle + \frac{L}{2} \|\mathbf{Y} - \mathbf{X}\|_F^2.$$

Both Conditions 4.3 and 4.4 can be derived for our specific examples discussed in Section 3.2. Based on Conditions 4.3 and 4.4, we prove that the sample loss function $\mathcal{L}_N$ satisfies a projected notion of the restricted Lipschitz continuous gradient property in Lemma C.1. It is worth noting that this new notion of restricted Lipschitz continuous gradient has not been appeared in the literature, and is essential to analyze the nonconvex optimization derived from low-rank matrix factorization. We believe it can be of broader interests for other non-convex problems to prove tight bounds.

Moreover, we assume that the gradient of the sample loss function $\nabla \mathcal{L}_N$ at $\mathbf{X}^*$ is upper bounded in terms of spectral norm.

**Condition 4.5.** Recall the unknown rank-$r$ matrix $\mathbf{X}^* \in \mathbb{R}^{d_1 \times d_2}$. Given a fixed sample size $N$ and tolerance parameter $\delta \in (0, 1)$, we let $\epsilon(N, \delta)$ be the smallest scalar such that with probability at least $1 - \delta$, we have

$$\|\nabla \mathcal{L}_N(\mathbf{X}^*)\|_2 \leq \epsilon(N, \delta),$$

where $\epsilon(N, \delta)$ depends on sample size $N$ and $\delta$.

Finally, we assume that each component loss function $\mathcal{L}_i$ defined in (3.5) satisfies the restricted strong smoothness condition.

**Condition 4.6** (Restricted Strong Smoothness for each Component). Given a fixed batch size $b$, assume $\mathcal{L}_i$ is restricted strongly smooth with parameter $L'$, such that for all matrices $\mathbf{X}, \mathbf{Y} \in \mathbb{R}^{d_1 \times d_2}$ with rank at most $3r$

$$\mathcal{L}_i(\mathbf{Y}) \leq \mathcal{L}_i(\mathbf{X}) + \langle \nabla \mathcal{L}_i(\mathbf{X}), \mathbf{Y} - \mathbf{X} \rangle + \frac{L'}{2} \|\mathbf{Y} - \mathbf{X}\|_F^2.$$

We will later show that for each illustrative example, the corresponding component loss function $\mathcal{L}_i$ indeed satisfies the restricted strong smoothness Condition 4.6.

## 4.1 Results for the Generic Setting

The following theorem shows that, in general, Algorithm 1 converges linearly to the unknown low-rank matrix $\mathbf{X}^*$ up to a statistical precision. We provide its proof in Section 5.1.

**Theorem 4.7** (SVRG). Suppose that Conditions 4.3, 4.4, 4.5, and 4.6 are satisfied. There exist constants $c_1, c_2, c_3$ and $c_4$ such that for any $\widetilde{\mathbf{Z}}^0 = [\widetilde{\mathbf{U}}^0; \widetilde{\mathbf{V}}^0] \in \mathbb{B}(c_2\sqrt{\sigma_r})$ with $c_2 \leq \min\{1/4, \sqrt{2\mu'/(5(3L+1))}\}$, if the sample size $N$ is large enough such that

$$\epsilon^2(N,\delta) \leq \frac{c_2^2(1-\rho)\mu'\sigma_r^2}{c_3 r},$$

where $\mu' = \min\{\mu, 1\}$, and the contraction parameter $\rho$ is defined as follows

$$\rho = \frac{10\kappa}{\mu'}\left(\frac{1}{\eta m \sigma_1} + c_4\eta\sigma_1 L'^2\right),$$

then with the step size $\eta = c_1/\sigma_1$ and the number of iterations $m$ properly chosen, the estimator $\widetilde{\mathbf{Z}}^S = [\widetilde{\mathbf{U}}^S; \widetilde{\mathbf{V}}^S]$ outputed from Algorithm 1 satisfies

$$\mathbb{E}[d^2(\widetilde{\mathbf{Z}}^S, \mathbf{Z}^*)] \leq \rho^S d^2(\widetilde{\mathbf{Z}}^0, \mathbf{Z}^*) + \frac{c_3 r \epsilon^2(N,\delta)}{(1-\rho)\mu'\sigma_r}, \quad (4.1)$$

with probability at least $1 - \delta$.

**Remark 4.8.** Theorem 4.7 implies that to achieve linear rate of convergence, it is necessary to set the step size $\eta$ to be small enough and the inner loop iterations $m$ to be large enough such that $\rho < 1$. Here we present a specific example to demonstrate such $\rho$ is attainable. As stated in Theorem 4.7, if we set the step size $\eta = c_1'/\sigma_1$, where $c_1' = \mu'/(15c_4\kappa L'^2)$, then the contraction parameter $\rho$ is calculated as follows

$$\rho = \frac{10\kappa}{m\mu'\eta\sigma_1} + \frac{2}{3}.$$

Therefore, under condition that $m \geq c_5\kappa^2$, we obtain $\rho \leq 5/6 < 1$, which leads to the linear convergence rate of Algorithm 1. Note that the step size $\eta$ is choosen as $c_1/\sigma_1$. While in practice, we can replace $\sigma_1$ by $\|\widetilde{\mathbf{Z}}^0\|_2^2$, since $\|\widetilde{\mathbf{Z}}^0\|_2 = O(\sqrt{\sigma_1})$. Besides, our algorithm also achieves the linear convergence in terms of reconstruction error, since the reconstruction error $\|\widetilde{\mathbf{X}}^s - \mathbf{X}^*\|_F^2$ can be upper bounded by $C\sigma_1 \cdot d^2(\widetilde{\mathbf{Z}}^s, \mathbf{Z}^*)$, where $C$ is a constant.

**Remark 4.9.** The right hand side of (4.1) consists of two parts, where the first part represents the optimization error and the second part denotes the statistical error. Note that in the noiseless case, since $\epsilon(N,\delta) = 0$, the statistical error becomes zero. As stated in Remark 4.8, with appropriate $\eta$ and $m$, we are able to achieve the linear rate of convergence. Therefore, in order to make sure the optimization error satisfies $\rho^S d^2(\widetilde{\mathbf{Z}}^0, \mathbf{Z}^*) \leq \epsilon$, it suffices to perform $S = O(\log(1/\epsilon))$ outer loop iterations. Recall that from Remark 4.8 we have $m = O(\kappa^2)$. Since for each outer loop iteration, it is required to calculate $m$ mixed stochastic variance-reduced gradients and one full gradient, the overall computational complexity for our algorithm to achieve $\epsilon$ precision is

$$O\left((Nd'^2 + \kappa^2 b d'^2)\log\left(\frac{1}{\epsilon}\right)\right),$$

where $d' = \max\{d_1, d_2\}$. However, the overall computational complexity of the state-of-the-art gradient descent based algorithm (Wang et al., 2016) to achieve $\epsilon$ precision is $O(N\kappa d'^2 \log(1/\epsilon))$. Therefore, provided that $\kappa \leq n$, our method is computationally more efficient than the state-of-the-art gradient descent approach. The detailed comparison of the overall computational complexity among different methods for each specific model can be found in next subsection.

The following theorem demonstrates the linear rate of convergence regarding our proposed initialization Algorithm 2. We refer Wang et al. (2016) to readers for a detailed proof.

**Theorem 4.10.** (Wang et al., 2016) Suppose the sample loss function $\mathcal{L}_N$ satisfies Conditions 4.3, 4.4 and 4.5. Let $\widetilde{\mathbf{X}}^0 = \widetilde{\mathbf{U}}^0 \widetilde{\mathbf{V}}^{0\top}$, where $(\widetilde{\mathbf{U}}^0, \widetilde{\mathbf{V}}^0)$ is the produced initial solution in Algorithm 2. If $L/\mu \in (1, 4/3)$, then with step size $\tau = 1/L$, we have

$$\|\widetilde{\mathbf{X}}^0 - \mathbf{X}^*\|_F \leq \rho^T \|\mathbf{X}^*\|_F + \frac{2\sqrt{3r}\epsilon(N, \delta)}{L(1-\rho)}, \tag{4.2}$$

with probability at least $1 - \delta$, where $\rho = 2\sqrt{1 - \mu/L}$ is the contraction parameter.

**Remark 4.11.** The right hand side of (4.2) again consists of two parts, where the first part denotes the optimization error and the second part represents the statistical error. To satisfy the initial assumption $\widetilde{\mathbf{Z}}^0 \in \mathbb{B}(c_2\sqrt{\sigma_r})$ in Theorem 4.7, it suffices to guarantee that $\widetilde{\mathbf{X}}^0$ is close enough to the unknown rank-$r$ matrix $\mathbf{X}^*$ such that $\|\widetilde{\mathbf{X}}^0 - \mathbf{X}^*\|_F \leq c\sigma_r$, where $c \leq \min\{1/2, 2c_2\}$. In fact, if $\|\widetilde{\mathbf{X}}^0 - \mathbf{X}^*\|_2 \leq c\sigma_r$, then according to Lemma D.1, we obtain

$$d^2(\widetilde{\mathbf{Z}}^0, \mathbf{Z}^*) \leq \frac{\sqrt{2} - 1}{2} \frac{\|\widetilde{\mathbf{X}}^0 - \mathbf{X}^*\|_F^2}{\sigma_r} \leq c_2^2 \sigma_r.$$

Therefore, in order to guarantee $\|\mathbf{X}^0 - \mathbf{X}^*\|_F \leq c\sigma_r$, based on (4.2) we need to perform at least $T = \log(c'\sigma_r/\|\mathbf{X}^*\|_F)/\log(\rho)$ number of iterations to ensure the optimization error is small enough, and it is also necessary to make sure the sample size $N$ is large enough such that

$$\epsilon(N, \delta) \leq \frac{c'L(1-\rho)\sigma_r}{2\sqrt{3r}},$$

which correponds to a sufficiently small statistical error.

## 4.2 Implications for Specific Models

In this subsection, we demonstrate the application of our generic theory to specific models. For each specific model, we only need to verify the restricted strong convexity and smoothness conditions 4.3, 4.4 for the corresponding sample loss function $\mathcal{L}_N$, the restricted smoothness condition 4.6 for each component loss function $\mathcal{L}_i$, and the statistical error bound condition 4.5. Here, we only lay out the main results for each model, the verification for these conditions can be found in Appendix A.

### 4.2.1 Matrix Sensing

We establish the theoretical guarantees of our algorithm for matrix sensing. First, for sample loss function $\mathcal{L}_N$, we establish the restricted strong convexity and smoothness conditions with parameters $\mu = 4/9$ and $L = 5/9$. Next, for each component loss function $\mathcal{L}_i$, we obtain the restricted strong smoothness parameter $L' = c_0 > 5/9$. Furthermore, we derive the upper bound of $\nabla \mathcal{L}_N(\mathbf{X}^*)$ in terms of spectral norm. If we choose the step size $\eta = c'_1/\sigma_1$, where $c'_1 = \mu'/(c'_0\kappa)$, and the inner loop iterations $m \geq c'_2\kappa^2$, where $c'_0, c'_1$ and $c'_2$ are some constants, then we have the following convergence result of our algorithm for the model of matrix sensing.

**Corollary 4.12.** Suppose the previously stated conditions are satisfied. For the estimator $\widetilde{\mathbf{Z}}^s$ produced at stage $s$ in Algorithm 1, if the initial solution satisfies $\widetilde{\mathbf{Z}}^0 \in \mathbb{B}(c_1\sqrt{\sigma_r})$, we have, with probability at least $1 - c_2 \exp(-c_3 d')$, that

$$\mathbb{E}\big[d^2(\widetilde{\mathbf{Z}}^S, \mathbf{Z}^*)\big] \leq \rho^S d^2(\widetilde{\mathbf{Z}}^0, \mathbf{Z}^*) + c_4 \nu^2 \frac{rd'}{N}, \quad (4.3)$$

where the contraction parameter $\rho < 1$, and $c_1, c_2, c_3, c_4$ are some constants.

**Remark 4.13.** The right hand side of (4.3) consists of two parts. The first one denotes the optimization error and the second one represents the statistical error. In particular, for the noisy case, the estimator produced by our algorithm achieves $O(\sqrt{rd'/N})$ statistical error after $O(\log(N/(rd')))$ number of outer loop iterations. This statistical error matches the minimax lower bound for matrix sensing (Negahban and Wainwright, 2011). And in the noiseless case, to ensure the restricted strong convexity and smoothness conditions of our objective function, we require sample size $N = O(rd')$, which attains the optimal sample complexity for matrix sensing (Recht et al., 2010; Tu et al., 2015; Wang et al., 2016). Most importantly, from Remark 4.9, we know that for the output $\widetilde{\mathbf{Z}}^S$ of our algorithm, the overall computational complexity of the estimator to achieve $\epsilon$ precision for matrix sensing is $O\big((Nd'^2 + \kappa^2 bd'^2)\log(1/\epsilon)\big)$. Nevertheless, the overall computational complexity for the state-of-the-art gradient descent algorithms for both noiseless (Tu et al., 2015) and noisy (Wang et al., 2016) cases to obtain $\epsilon$ precision is $O\big(N\kappa d'^2 \log(1/\epsilon)\big)$. Our approach is more efficient under the condition that $\kappa \leq n$. And this result is consistent with the result obtained by Zhang et al. (2017). In their work, they proposed an accelerated stochastic gradient descent method for matrix sensing based on the restricted isometry property. Nevertheless, since the restricted isometry property is more restrictive than the restricted strong convex and smoothness conditions, their results can not be applied to general low-rank matrix recovery problems, like one-bit matrix completion.

### 4.2.2 Matrix Completion

We establish the theoretical guarantees of our algorithm for matrix completion. In particular, we consider a partial observation model, which means only the elements over a subset $\mathcal{X} \subseteq [d_1] \times [d_2]$ are observed. In addition, we assume a uniform sampling model for $\mathcal{X}$, which is defined as

$$\forall (j, k) \in \mathcal{X}, \ j \sim \text{uniform}([d_1]), \ k \sim \text{uniform}([d_2]).$$

To establish the restricted strong convexity and smoothness conditions for $\mathcal{L}_N$ in matrix completion, we impose an infinity norm constraint on our estimator $\|\mathbf{X}\|_\infty \leq \alpha$, which is also known as spikiness condition (Negahban and Wainwright, 2012). In addition, this condition is argued to be much less restrictive than the incoherence conditions (Candès and Recht, 2009) assumed in exact low-rank matrix completion (Negahban and Wainwright, 2012; Klopp et al., 2014). The more detailed discussion of this condition can be found in section A.2.

Let $\mathcal{C}(\alpha) = \big\{\mathbf{X} \in \mathbb{R}^{d_1 \times d_2} \,\big|\, \|\mathbf{X}\|_\infty \leq \alpha\big\}$ be the infinity norm constraint set. In order to make sure our estimator can satisfy this constraint, we need a projection step, which is displayed in Algorithm

1. Therefore, we construct two feasible sets $\mathcal{C}_i = \{\mathbf{A} \in \mathbb{R}^{d_i \times r} \mid \|\mathbf{A}\|_{2,\infty} \leq \sqrt{\alpha}\}$, where $i \in \{1, 2\}$, based on $\mathcal{C}(\alpha)$. Thus, for any $\mathbf{U} \in \mathcal{C}_1$ and $\mathbf{V} \in \mathcal{C}_2$, we can ensure $\mathbf{X} = \mathbf{U}\mathbf{V}^\top \in \mathcal{C}(\alpha)$.

By assuming spikiness condition in matrix completion, we can obtain the restricted strongly convex and smooth conditions for $\mathcal{L}_N$ with parameters $\mu = 8/9$ and $L = 10/9$. In addition, for each component loss function $\mathcal{L}_i$, we can get the restricted strong smoothness parameter $L' = c_0 > 10/9$. Furthermore, we can derive the upper bound of $\nabla \mathcal{L}_N(\mathbf{X}^*)$ in terms of spectral norm. If we choose the step size $\eta = c_1'/\sigma_1$, where $c_1' = \mu'/(c_0'\kappa)$, and the inner loop iterations $m \geq c_2'\kappa^2$, where $c_0', c_1'$ and $c_2'$ are some constants, then we have the following convergence result of our algorithm for the model of matrix completion.

**Corollary 4.14.** Suppose the previously stated conditions are satisfied and $\mathbf{X}^* \in \mathcal{C}(\alpha)$. For the estimator $\widetilde{\mathbf{Z}}^s$ produced at stage $s$ in Algorithm 1, if the initial solution satisfies $\widetilde{\mathbf{Z}}^0 \in \mathbb{B}(c_1\sqrt{\sigma_r})$, we have, with probability at least $1 - c_2/d'$, that

$$\mathbb{E}\big[d^2(\widetilde{\mathbf{Z}}^S, \mathbf{Z}^*)\big] \leq \rho^S d^2(\widetilde{\mathbf{Z}}^0, \mathbf{Z}^*) + c_3 \max\{\nu^2, \alpha^2\} \frac{rd' \log d'}{p}, \tag{4.4}$$

where the contraction parameter $\rho < 1$, and $c_1, c_2, c_3$ are some constants.

**Remark 4.15.** In the noisy case, for the standardized error term $\|\mathbf{X}^S - \mathbf{X}^*\|_F/\sqrt{d_1 d_2}$, our algorithm achieves $O\big(\sqrt{rd' \log d'/N}\big)$ statistical error after $O\big(\log(N/(rd' \log d'))\big)$ number of outer loop iterations. This statistical error matches the minimax lower bound for matrix completion provided by Negahban and Wainwright (2012); Koltchinskii et al. (2011). And in the noiseless case, to ensure the restricted strong convexity and smoothness conditions of our objective function, we require the sample size $N = O(rd' \log d')$, which attains optimal sample complexity (Candès and Recht, 2009; Recht, 2011; Chen et al., 2013). Recall that from Remark 4.9, the overall computational complexity of our algorithm to reach $\epsilon$ accuracy for matrix completion is $O\big((Nd'^2 + \kappa^2 bd'^2) \log(1/\epsilon)\big)$. However, for the state-of-the-art gradient descent based algorithms, the computational complexity for both noiseless (Zheng and Lafferty, 2016) and noisy (Wang et al., 2016) settings to obtain $\epsilon$ accuracy is $O\big(N\kappa d'^2 \log(1/\epsilon)\big)$. Thus the computational complexity of our algorithm is lower than the state-of-the-art gradient descent methods if we have $\kappa \leq n$. In addition, for the online stochastic gradient descent algorithm (Jin et al., 2016), the overall computational complexity is $O(r^4 d' \kappa^4 \log(1/\epsilon))$. Since their results has a forth power dependency on both $r$ and $\kappa$, our method can yield a significant improvement over the online method when $r, \kappa$ is large. Another drawback of their method is that they require $O\big(rd'\kappa^4(r + \log(1/\epsilon)) \log d'\big)$ sample complexity, which is much higher than ours $O(rd' \log d')$.

### 4.2.3 One-bit Matrix Completion

We establish the theoretical guarantees of our algorithm for one-bit matrix completion. We obtain the restricted strong convexity and smoothness conditions for $\mathcal{L}_N$ with parameters $\mu = C_1 \mu_\alpha$ and $L = C_2 L_\alpha$. In addition, we are able to get the restricted strong smoothness condition for each component function $\mathcal{L}_i$ with parameter $L' = c_0 L_\alpha > L$. Here, $\mu_\alpha$ and $L_\alpha$ are defined as

$$\mu_\alpha \leq \min\left(\inf_{|x|\leq \alpha}\left\{\frac{f'^2(x)}{f^2(x)} - \frac{f''(x)}{f(x)}\right\}, \inf_{|x|\leq \alpha}\left\{\frac{f'^2(x)}{(1-f(x))^2} + \frac{f''(x)}{1-f(x)}\right\}\right), \tag{4.5}$$

$$L_\alpha \geq \max\left(\sup_{|x|\leq \alpha}\left\{\frac{f'^2(x)}{f^2(x)} - \frac{f''(x)}{f(x)}\right\}, \sup_{|x|\leq \alpha}\left\{\frac{f'^2(x)}{(1-f(x))^2} + \frac{f''(x)}{1-f(x)}\right\}\right), \tag{4.6}$$

where $f(x)$ is the probability density function, and each element $X_{jk}$ satisfies $|X_{jk}| \leq \alpha$. Note that given the probability density function $f(x)$ and constant $\alpha$, we can calculate $\mu_\alpha$ and $L_\alpha$, which are fixed constants and do not rely on dimension of the unknown low-rank matrix. For example, if we have logistic function as the probability density function, we can get $\mu_\alpha = e^\alpha/(1+e^\alpha)^2$ and $L_\alpha = 1/4$. Furthermore, we define $\gamma_\alpha$ as follows, which reflects the steepness property of the sample loss function $\mathcal{L}_N(\cdot)$

$$\gamma_\alpha \geq \sup_{|x| \leq \alpha} \left\{ \frac{|f'(x)|}{f(x)(1-f(x))} \right\}. \tag{4.7}$$

Moreover, we can derive the upper bound of the $\nabla \mathcal{L}_N(\mathbf{X}^*)$ in terms of spectral norm. If we choose the step size $\eta = c_1'/\sigma_1$, where $c_1' = \mu'/(c_0'\kappa)$, and the inner loop iterations $m \geq c_2'\kappa^2$, where $c_0', c_1'$ and $c_2'$ are some constants, then we have the following convergence result of our algorithm for the model of matrix completion.

**Corollary 4.16.** Suppose the previously stated conditions are satisfied and $\mathbf{X}^* \in \mathcal{C}(\alpha)$. Furthermore, suppose only a subset of the unknown low-rank matrix $\mathbf{X}^*$ is observed, and index set $\Omega$ is sampled from uniform sampling model. Given the binary observation matrix $\mathbf{Y}$ in (3.6) with log-concave probability density function $f$, for the estimator $\widetilde{\mathbf{Z}}^s$ produced at stage $s$ in Algorithm 1, if the initial solution $\widetilde{\mathbf{Z}}^0 \in \mathbb{B}(c_1\sqrt{\sigma_r})$, we have, with probability at least $1 - c_2/d'$, that

$$\mathbb{E}\big[d^2(\widetilde{\mathbf{Z}}^S, \mathbf{Z}^*)\big] \leq \rho^S d^2(\widetilde{\mathbf{Z}}^0, \mathbf{Z}^*) + c_3 \max\{\gamma_\alpha^2, \alpha^2\} \frac{rd'\log d'}{p}, \tag{4.8}$$

where the contraction parameter $\rho < 1$, and $c_1, c_2, c_3$ are some constants.

**Remark 4.17.** For the standardized error term $\|\mathbf{X}^S - \mathbf{X}^*\|_F/\sqrt{d_1 d_2}$, our algorithm achieves $O\big(\sqrt{rd'\log d'/N}\big)$ statistical error after $O\big(\log(N/(rd'\log d'))\big)$ number of outer loop iterations. This statistical error matches the minimax lower bound of one-bit matrix completion established in Davenport et al. (2014); Cai and Zhou (2013). Furthermore, Remark 4.9 tells us that for our estimator $\widetilde{\mathbf{Z}}^S$ to achieve $\epsilon$ accuracy, the overall computational complexity required by our algorithm is $O\big((Nd'^2 + \kappa^2 bd'^2)\log(1/\epsilon)\big)$. Nevertheless, the overall computational complexity for the state-of-the-art gradient descent based algorithm (Wang et al., 2016) to obtain $\epsilon$ accuracy is $O\big(N\kappa d'^2 \log(1/\epsilon)\big)$. Therefore, as long as we have $\kappa \leq n$, our approach is more efficient than the state-of-the-art gradient descent method. Moreover, the overall computational complexities for the state-of-the-art projected gradient descent algorithm (Chen and Wainwright, 2015) and the conditional gradient descent (a.k.a., Frank-Wolfe) algorithm (Ni and Gu, 2016) to obtain $\epsilon$ accuracy are both $O\big(Nrd'^2 \log(1/\epsilon)\big)$[1]. If we have $\kappa^2 \leq nr$, our method clearly has a lower computational complexity than theirs.

---

[1] Note that the overall computational complexities for the projected gradient descent (Chen and Wainwright, 2015) and conditional gradient descent (Ni and Gu, 2016) algorithms also depend on some problem dependent parameters, which we omit here but actually can make their computational complexities worse. Please refer to their papers for more accurate complexity results.

# 5 Proof of the Main Theory

We provide the proof of our main theoretical results in this section. According to (3.3), we aim to minimize the following objective function in terms of $\mathbf{Z} = [\mathbf{U}; \mathbf{V}]$

$$\widetilde{F}_N(\mathbf{Z}) = F_N(\mathbf{U}, \mathbf{V}) = \mathcal{L}_N(\mathbf{U}\mathbf{V}^\top) + \frac{1}{8}\|\mathbf{U}^\top\mathbf{U} - \mathbf{V}^\top\mathbf{V}\|_F^2. \tag{5.1}$$

Therefore, we obtain the corresponding gradient

$$\nabla \widetilde{F}_N(\mathbf{Z}) = \begin{bmatrix} \nabla_\mathbf{U}\mathcal{L}_N(\mathbf{U}\mathbf{V}^\top) + \frac{1}{2}\mathbf{U}(\mathbf{U}^\top\mathbf{U} - \mathbf{V}^\top\mathbf{V}) \\ \nabla_\mathbf{V}\mathcal{L}_N(\mathbf{U}\mathbf{V}^\top) + \frac{1}{2}\mathbf{V}(\mathbf{U}^\top\mathbf{U} - \mathbf{V}^\top\mathbf{V}) \end{bmatrix}. \tag{5.2}$$

## 5.1 Proof of Theorem 4.7

In order to prove Theorem 4.7, we need following lemmas, and we present the corresponding proofs in Sections B.1 and B.2, respectively.

**Lemma 5.1** (Local Curvature Condition). Suppose the sample loss function $\mathcal{L}_N$ satisfies Conditions 4.3 and 4.4. For any matrix $\mathbf{Z} = [\mathbf{U}; \mathbf{V}] \in \mathbb{R}^{(d_1+d_2)\times r}$, where $\mathbf{U} \in \mathbb{R}^{d_1 \times r}$ and $\mathbf{V} \in \mathbb{R}^{d_2 \times r}$, denote $\widetilde{\mathbf{Z}} = [\mathbf{U}; -\mathbf{V}]$. In addition, we use $\mathbf{R} = \operatorname{argmin}_{\widetilde{\mathbf{R}} \in \mathbb{Q}_r} \|\mathbf{Z} - \mathbf{Z}^*\widetilde{\mathbf{R}}\|_F$ to denote the optimal rotation with respect to $\mathbf{Z}$, and $\mathbf{H} = \mathbf{Z} - \mathbf{Z}^*\mathbf{R}$, then the following inequality holds

$$\langle \nabla \widetilde{F}_N(\mathbf{Z}), \mathbf{H}\rangle \geq \frac{\mu}{8}\|\mathbf{X} - \mathbf{X}^*\|_F^2 + \frac{\mu'\sigma_r}{10}\|\mathbf{H}\|_F^2 + \frac{1}{16}\|\widetilde{\mathbf{Z}}^\top\mathbf{Z}\|_F^2 \\ - \frac{3L+1}{8}\|\mathbf{H}\|_F^4 - \left(\frac{4r}{\mu} + \frac{r}{2L}\right)\cdot\|\nabla\mathcal{L}_N(\mathbf{X}^*)\|_2^2,$$

where $\mathbf{X} = \mathbf{U}\mathbf{V}^\top$, and $\mu' = \min\{\mu, 1\}$.

**Lemma 5.2** (Local Smoothness Condition). Assume the component loss function $\mathcal{L}_i$ satisfies Condition 4.6. Suppose we randomly pick $i \in [n]$. For any $\mathbf{U} \in \mathbb{R}^{d_1 \times r}$, $\mathbf{V} \in \mathbb{R}^{d_2 \times r}$ and rank-$r$ matrix $\widetilde{\mathbf{X}} \in \mathbb{R}^{d_1 \times d_2}$, we denote $\mathbf{Z} = [\mathbf{U}; \mathbf{V}]$ and $\mathbf{X} = \mathbf{U}\mathbf{V}^\top$. Let $\mathbf{G}_U = \nabla_\mathbf{U} F_i(\mathbf{U}, \mathbf{V}) - \nabla \mathcal{L}_i(\widetilde{\mathbf{X}})\mathbf{V} + \nabla\mathcal{L}_N(\widetilde{\mathbf{X}})\mathbf{V}$, $\mathbf{G}_V^t = \nabla_\mathbf{V} F_i(\mathbf{U}, \mathbf{V}) - \nabla\mathcal{L}_i(\widetilde{\mathbf{X}})^\top\mathbf{U} + \nabla\mathcal{L}_N(\mathbf{X})^\top\mathbf{U}$, and $\mathbf{G} = [\mathbf{G}_U; \mathbf{G}_V]$. Then we have

$$\mathbb{E}\|\mathbf{G}\|_F^2 \leq 24\bigl(2L'^2\|\widetilde{\mathbf{X}} - \mathbf{X}^*\|_F^2 + (2L'^2 + L^2)\cdot\|\mathbf{X} - \mathbf{X}^*\|_F^2\bigr)\cdot\|\mathbf{Z}\|_2^2 \\ + \|\mathbf{U}^\top\mathbf{U} - \mathbf{V}^\top\mathbf{V}\|_F^2 \cdot \|\mathbf{Z}\|_2^2 + 12r\|\nabla\mathcal{L}_N(\mathbf{X}^*)\|_2^2 \cdot \|\mathbf{Z}\|_2^2.$$

*Proof of Theorem 4.7.* According to stochastic variance reduced gradient descent Algorithm 1, consider iteration $t$ in the inner loop, we have the following update

$$\mathbf{U}^{t+1} = \mathcal{P}_{\mathcal{C}_1}(\mathbf{U}^t - \eta \mathbf{G}_U^t), \text{ and } \mathbf{V}^{t+1} = \mathcal{P}_{\mathcal{C}_2}(\mathbf{V}^t - \eta \mathbf{G}_V^t),$$

where we denote

$$\mathbf{G}_U^t = \nabla_\mathbf{U} F_{i_t}(\mathbf{U}^t, \mathbf{V}^t) - \nabla\mathcal{L}_{i_t}(\widetilde{\mathbf{X}})\mathbf{V}^t + \nabla\mathcal{L}_N(\widetilde{\mathbf{X}})\mathbf{V}^t, \\ \mathbf{G}_V^t = \nabla_\mathbf{V} F_{i_t}(\mathbf{U}^t, \mathbf{V}^t) - \nabla\mathcal{L}_{i_t}(\widetilde{\mathbf{X}})^\top\mathbf{U}^t + \nabla\mathcal{L}_N(\mathbf{X}^t)^\top\mathbf{U}^t.$$

Since $i_t$ is uniformly picked from $[n]$, we have $\mathbb{E}[\mathbf{G}_U^t] = \nabla_\mathbf{U} F_N(\mathbf{U}^t, \mathbf{V}^t)$ and $\mathbb{E}[\mathbf{G}_V^t] = \nabla_\mathbf{V} F_N(\mathbf{U}^t, \mathbf{V}^t)$, where the expectation is taken with respect to $i_t$. Recall $\mathbf{Z}^t = [\mathbf{U}^t; \mathbf{V}^t]$, and $\mathbf{R}^t = \operatorname{argmin}_{\mathbf{R}\in\mathbb{Q}_r}\|\mathbf{Z}^t - $ $\mathbf{Z}^*\mathbf{R}\|_F$ as the optimal rotation with respect to $\mathbf{Z}^t$. Denote $\mathbf{H}^t = \mathbf{Z} - \mathbf{Z}^*\mathbf{R}^t$ and $\mathbf{G}^t = [\mathbf{G}^t_U; \mathbf{G}^t_V]$. By induction, for any $t \geq 0$, we assume $\mathbf{Z}^t \in \mathbb{B}(c_2\sqrt{\sigma_r})$. Thus, by taking the expectation of $\mathbf{H}^{t+1}$ over $i_t$ conditioned on $\mathbf{Z}^t$, we have

$$\begin{aligned}
\mathbb{E}\|\mathbf{H}^{t+1}\|_F^2 &\leq \mathbb{E}\|\mathcal{P}_{\mathcal{C}_1}(\mathbf{U}^t - \eta\mathbf{G}^t_U) - \mathbf{U}^*\mathbf{R}^t\|_F^2 + \mathbb{E}\|\mathcal{P}_{\mathcal{C}_2}(\mathbf{V}^t - \eta\mathbf{G}^t_V) - \mathbf{V}^*\mathbf{R}^t\|_F^2 \\
&\leq \mathbb{E}\|\mathbf{U}^t - \eta\mathbf{G}^t_U - \mathbf{U}^*\mathbf{R}^t\|_F^2 + \mathbb{E}\|\mathbf{V}^t - \eta\mathbf{G}^t_V - \mathbf{V}^*\mathbf{R}^t\|_F^2 \\
&= \|\mathbf{H}^t\|_F^2 - 2\eta\mathbb{E}\langle\mathbf{G}^t, \mathbf{H}^t\rangle + \eta^2\mathbb{E}\|\mathbf{G}^t\|_F^2 \\
&= \|\mathbf{H}^t\|_F^2 - 2\eta\langle\nabla\widetilde{F}_N(\mathbf{Z}^t), \mathbf{H}^t\rangle + \eta^2\mathbb{E}\|\mathbf{G}^t\|_F^2, \quad (5.3)
\end{aligned}$$

where the first inequality follows from the definition of $\mathbf{H}^t$, the second inequality follows from the non-expansive property of the projection $\mathcal{P}_{\mathcal{C}_i}$ onto $\mathcal{C}_i$ and the fact that $\mathbf{U}^* \in \mathcal{C}_1, \mathbf{V}^* \in \mathcal{C}_2$, and the last equality holds because conditioned on $\mathbf{Z}^t$, $\mathbb{E}\langle\mathbf{H}^t, \mathbf{G}^t\rangle = \langle\mathbf{H}^t, \mathbb{E}\mathbf{G}^t\rangle = \langle\mathbf{H}^t, \nabla\widetilde{F}_N(\mathbf{Z}^t)\rangle$, where $\widetilde{F}_N$ is defined in (5.1). According to Lemma 5.1, we can obtain the lower bound of $\langle\nabla\widetilde{F}_N(\mathbf{Z}^t), \mathbf{H}^t\rangle$.

$$\begin{aligned}
\langle\nabla\widetilde{F}_N(\mathbf{Z}^t), \mathbf{H}^t\rangle &\geq \frac{\mu}{8}\|\mathbf{X}^t - \mathbf{X}^*\|_F^2 + \frac{\mu'\sigma_r}{10}\|\mathbf{H}^t\|_F^2 + \frac{1}{16}\|\mathbf{U}^{t\top}\mathbf{U}^t - \mathbf{V}^{t\top}\mathbf{V}^t\|_F^2 - \frac{3L+1}{8}\|\mathbf{H}^t\|_F^4 \\
&\quad - \left(\frac{4r}{\mu} + \frac{r}{2L}\right) \cdot \|\nabla\mathcal{L}_N(\mathbf{X}^*)\|_2^2, \quad (5.4)
\end{aligned}$$

where $\mu' = \min\{\mu, 1\}$. According to Lemma 5.2, we have

$$\begin{aligned}
\mathbb{E}\|\mathbf{G}^t\|_F^2 &\leq 24\big(2L'^2\|\widetilde{\mathbf{X}}^t - \mathbf{X}^*\|_F^2 + (2L'^2 + L^2) \cdot \|\mathbf{X}^t - \mathbf{X}^*\|_F^2\big) \cdot \|\mathbf{Z}^t\|_2^2 \\
&\quad + \|\mathbf{U}^{t\top}\mathbf{U}^t - \mathbf{V}^{t\top}\mathbf{V}^t\|_F^2 \cdot \|\mathbf{Z}^t\|_2^2 + 12r\|\nabla\mathcal{L}_N(\mathbf{X}^*)\|_2^2 \cdot \|\mathbf{Z}^t\|_2^2. \quad (5.5)
\end{aligned}$$

Note that for any $\mathbf{Z} \in \mathbb{B}(\sqrt{\sigma_r}/4)$, denote $\mathbf{R}$ as the optimal rotation with respect to $\mathbf{Z}$, we have $\|\mathbf{Z}\|_2 \leq \|\mathbf{Z}^*\|_2 + \|\mathbf{Z} - \mathbf{Z}^*\mathbf{R}\|_2 \leq 2\sqrt{\sigma_1}$. Thus, we have $\|\mathbf{Z}^t\|_2^2 \leq 4\sigma_1$. Denote $L_m = \max\{L, L'\}$, and we let $\eta = c_1/\sigma_1$, where $c_1 \leq \min\{1/32, \mu/(1152L_m^2)\}$. Therefore, combining (5.4) and (5.5), we have

$$\begin{aligned}
-2\eta\langle\nabla\widetilde{F}_N(\mathbf{Z}), \mathbf{H}\rangle + \eta^2\mathbb{E}\|\mathbf{G}^t\|_F^2 &\leq -\frac{\eta\mu'\sigma_r}{5}\|\mathbf{H}^t\|_F^2 + \frac{\eta(3L+1)}{4}\|\mathbf{H}^t\|_F^4 + 192\eta^2\sigma_1 L'^2\|\widetilde{\mathbf{X}}^t - \mathbf{X}^*\|_F^2 \\
&\quad + \eta\left(\frac{8r}{\mu} + \frac{r}{L}\right) \cdot \|\nabla\mathcal{L}_N(\mathbf{X}^*)\|_2^2 + 48\eta^2\sigma_1 r\|\nabla\mathcal{L}_N(\mathbf{X}^*)\|_2^2.
\end{aligned}$$

Note that according to our assumption, $\|\mathbf{H}^t\|_F^2 \leq c_2^2\sigma_r$ with $c_2^2 \leq 2\mu'/(5(3L+1))$. Thus, according to Condition 4.5, we further have

$$-2\eta\langle\nabla\widetilde{F}_N(\mathbf{Z}), \mathbf{H}\rangle + \eta^2\mathbb{E}\|\mathbf{G}^t\|_F^2 \leq -\frac{\eta\mu'\sigma_r}{10}\|\mathbf{H}^t\|_F^2 + 192\eta^2\sigma_1 L'^2\|\widetilde{\mathbf{X}}^t - \mathbf{X}^*\|_F^2 + c_3\eta r\epsilon^2(N, \delta), \quad (5.6)$$

holds with probability at least $1 - \delta$, where $c_3 \geq 48c_1 + 8/\mu + 1/L$. Therefore, plugging (5.6) into (5.3), with probability at least $1 - \delta$, we have

$$\mathbb{E}\|\mathbf{H}^{t+1}\|_F^2 \leq \left(1 - \frac{\eta\mu'\sigma_r}{10}\right) \cdot \|\mathbf{H}^t\|_F^2 + 192\eta^2\sigma_1 L'^2\|\widetilde{\mathbf{X}}^t - \mathbf{X}^*\|_F^2 + c_3\eta r\epsilon^2(N, \delta). \quad (5.7)$$

Finally, for a fixed stage of $s$, we have $\widetilde{\mathbf{X}} = \widetilde{\mathbf{X}}^{s-1}$ accordingly. Denote $\widetilde{\mathbf{Z}}^s = [\widetilde{\mathbf{U}}^s; \widetilde{\mathbf{V}}^s]$, for any $s$. According to Algorithm 1, we randomly choose $\widetilde{\mathbf{Z}}^s$ after all of the updates are completed. Therefore, we first take summation of the previous inequality (5.7) over $t \in \{0, 1, \cdots, m-1\}$, and then take expectation with regard to all the history, we can get

$$\mathbb{E}\|\mathbf{H}^m\|_F^2 - \mathbb{E}\|\mathbf{H}^0\|_F^2 \leq -\frac{\eta\mu'\sigma_r}{10}\sum_{t=0}^{m-1}\mathbb{E}\|\mathbf{H}^t\|_F^2 + 192\eta^2\sigma_1 L'^2 m\mathbb{E}\|\widetilde{\mathbf{X}}^{s-1} - \mathbf{X}^*\|_F^2 + c_3\eta mr\epsilon^2(N,\delta).$$

For any $s$, we denote $\widetilde{\mathbf{R}}^s = \operatorname{argmin}_{\mathbf{R} \in \mathbb{Q}_r}\|\widetilde{\mathbf{Z}}^s - \mathbf{Z}^*\mathbf{R}\|_F$ and $\widetilde{\mathbf{H}}^s = \widetilde{\mathbf{Z}}^s - \mathbf{Z}^*\widetilde{\mathbf{R}}^s$. According to the choice of $\widetilde{\mathbf{Z}}^s$ in Algorithm 1, we have

$$\mathbb{E}\|\widetilde{\mathbf{H}}^s\|_F^2 = \frac{1}{m}\sum_{t=0}^{m-1}\mathbb{E}\|\mathbf{H}^t\|_F^2.$$

Note that according to Algorithm 1, we have $\mathbf{H}^0 = \widetilde{\mathbf{H}}^{s-1}$, thus we further obtain

$$\mathbb{E}\|\mathbf{H}^m\|_F^2 - \mathbb{E}\|\widetilde{\mathbf{H}}^{s-1}\|_F^2 \leq -\frac{\eta m\mu'\sigma_r}{10}\mathbb{E}\|\widetilde{\mathbf{H}}^s\|_F^2 + 192\eta^2\sigma_1 L'^2 m\mathbb{E}\|\widetilde{\mathbf{X}}^{s-1} - \mathbf{X}^*\|_F^2 + c_3\eta mr\epsilon^2(N,\delta).$$

Note that $\widetilde{\mathbf{Z}}^{s-1} \in \mathbb{B}(\sqrt{\sigma_r}/4)$, thus according to Lemma D.3, we have

$$\|\widetilde{\mathbf{X}}^{s-1} - \mathbf{X}^*\|_F^2 \leq 3\|\mathbf{Z}^*\|_2^2 \cdot d^2(\widetilde{\mathbf{Z}}^{s-1}, \mathbf{Z}^*) = 6\sigma_1\|\widetilde{\mathbf{H}}^{s-1}\|_F^2.$$

where the first inequality follows from Lemma D.3 and the second inequality holds because $\|\mathbf{Z}^*\|_2 = \sqrt{2\sigma_r}$. Therefore, we obtain

$$\frac{\eta m\mu'\sigma_r}{10}\mathbb{E}\|\widetilde{\mathbf{H}}^s\|_F^2 \leq (1152\eta^2\sigma_1^2 L'^2 m + 1) \cdot \mathbb{E}\|\widetilde{\mathbf{H}}^{s-1}\|_F^2 + c_3\eta mr\epsilon^2(N,\delta),$$

holds with probability at least $1 - \delta$, which gives us following contraction parameter

$$\rho = \frac{10\kappa}{\mu'}\bigg(\frac{1}{\eta m\sigma_1} + 1152\eta L'^2\bigg).$$

Note that $\eta = c_1/\sigma_1$, hence we can let $\rho \in (0, 1)$ by choosing sufficiently small constant $c_1$ and sufficiently large number of iterations $m$. Therefore, with probability at least $1 - \delta$, we can get

$$\mathbb{E}\|\widetilde{\mathbf{H}}^s\|_F^2 \leq \rho\mathbb{E}\|\widetilde{\mathbf{H}}^{s-1}\|_F^2 + \frac{10c_3}{\mu'\sigma_r} \cdot r\epsilon^2(N,\delta).$$

$\square$

## 6 Experiments

In this section, we provide experimental results for specific models to illustrate the superiority of our proposed method. We first present the results for the convergence rate of the proposed stochastic variance reduced gradient descent algorithm and the state-of-the-art gradient descent algorithm (Wang et al., 2016) to demonstrate the effectiveness of our method. Note that both methods use the same initialization algorithm (Algorithm 2) originally proposed in Wang et al. (2016). Then we evaluate the sample complexity that is required for both methods to exactly recover the unknown low-rank matrix in the noiseless case. Finally, we investigate the statistical error of our method in the noisy case. Note that both algorithms use the same initialization method (Algorithm 2), and for the proposed stochastic variance reduced gradient descent algorithm, all the experimental results are based upon the optimal batch size $b$, iteration number $m$ and step size $\eta$, which are selected by cross validation, and averaged over 30 trails.

## 6.1 Matrix Sensing

For matrix sensing, we consider the unknown low-rank matrix $\mathbf{X}^*$ in the following settings: (i) $d_1 = 100, d_2 = 80, r = 2$; (ii) $d_1 = 120, d_2 = 100, r = 3$; (iii) $d_1 = 140, d_2 = 120, r = 4$. First, we generate the unknown low-rank matrix $\mathbf{X}^*$ as $\mathbf{X}^* = \mathbf{U}^*\mathbf{V}^{*\top}$, where $\mathbf{U}^* \in \mathbb{R}^{d_1 \times r}$ and $\mathbf{V}^* \in \mathbb{R}^{d_2 \times r}$ are randomly generated. Next, we obtain linear measurements from the following observation model $y_i = \langle \mathbf{A}_i, \mathbf{X}^* \rangle + \epsilon_i$, where each elements of the sensing matrix $\mathbf{A}_i$ follows i.i.d. standard normal distribution. Finally, we consider two settings: (1) noisy case: the noise follows i.i.d. normal distribution with variance $\sigma^2 = 0.25$ and (2) noiseless case.

For the results of the convergence rate, Figure 1(a) and 1(c) illustrate the squared relative error $\|\widehat{\mathbf{X}} - \mathbf{X}^*\|_F^2/\|\mathbf{X}^*\|_F^2$ in log scale versus number of effective data passes for both methods under setting (i). These results show the linear convergence rate of our method. Most importantly, it clearly demonstrates the superiority of our approach, since our algorithm shows better performance after the same number of effective data passes compared with the state-of-the-art gradient descent algorithm. Since we get results with similar patterns for other settings, we leave them out for simplicity. For the results of sample complexity, we illustrate the empirical probability of exact recovery under rescaled sample size $N/(rd')$. For the estimator $\widehat{\mathbf{X}}$ given by different algorithms, it is considered to be exact recovery, if the relative error $\|\widehat{\mathbf{X}} - \mathbf{X}^*\|_F/\|\mathbf{X}^*\|_F$ is less than $10^{-3}$. Figure 1(b) shows the empirical recovery probability of different methods under setting (i). The result implies a phase transition around $N = 3rd'$, which is consistent with the optimal sample complexity that $N$ is linear with $rd$. Besides, since we get results with similar patterns for other settings, we leave them out to save space. For the results of statistical error, Figure 1(d) shows, in the noisy case, how the estimation errors scale with the rescaled sample size $N/(rd')$, which confirms our theoretical results.

## 6.2 Matrix Completion

For matrix completion, we use the same procedure as in matrix sensing to generate $\mathbf{X}^*$, and we use uniform observation model to get observation matrix $\mathbf{Y}$. We also consider the same noisy and noiseless settings as in matrix sensing. For the results of convergence rate, we show the logarithm of mean squared error in terms of Frobenius norm $\|\widehat{\mathbf{X}} - \mathbf{X}^*\|_F^2/(d_1 d_2)$ versus number of effective data passes. Figure 2(a) and 2(c) illustrate the linear rate of convergence of our method under setting (i). In addition, the results imply that after the same number of effective data passes, our algorithm is more efficient than the state-of-the-art gradient descent based algorithm in estimation error. Figure 2(b) demonstrates the results of empirical exact recovery probability versus rescaled sample size $N/(rd' \log d')$ under setting (i). It implies a phase transition around $N = 3rd' \log d'$, which corroborate the optimal sample complexity $N = O(rd' \log d')$. Besides, we leave out results under other settings for simplicity since we get similar patterns for these results. For the results of statistical error, The results of statistical error are displayed in Figure 2(d), which is consistent with our Corollary 4.14.

## 6.3 One-bit Matrix Completion

We use the same settings of $\mathbf{X}^*$ for one-bit matrix completion as before. In order to obtain $\mathbf{X}^*$, we adopt the similar procedure as in Davenport et al. (2014); Bhaskar and Javanmard (2015); Ni and Gu (2016). In detail, we first randomly generate $\mathbf{U}^* \in \mathbb{R}^{d_1 \times r}, \mathbf{V}^* \in \mathbb{R}^{d_2 \times r}$ from a uniform distribution on $[-1/2, 1/2]$. Then we get $\mathbf{X}^*$ by $\mathbf{X}^* = \mathbf{U}^*\mathbf{V}^{*\top}$. Finally, we scale $\mathbf{X}^*$ to make it satisfies $\|\mathbf{X}^*\|_\infty = \alpha = 1$. Here we consider the uniform observation model with probability density function $f(X_{ij}) = \Phi(\mathbf{X}_{ij}/\sigma)$ in (3.6), where $\Phi$ is the cumulative density function of the standard normal distribution, and $\sigma$ is the noise level, which we set it to be $\sigma = 0.5$.

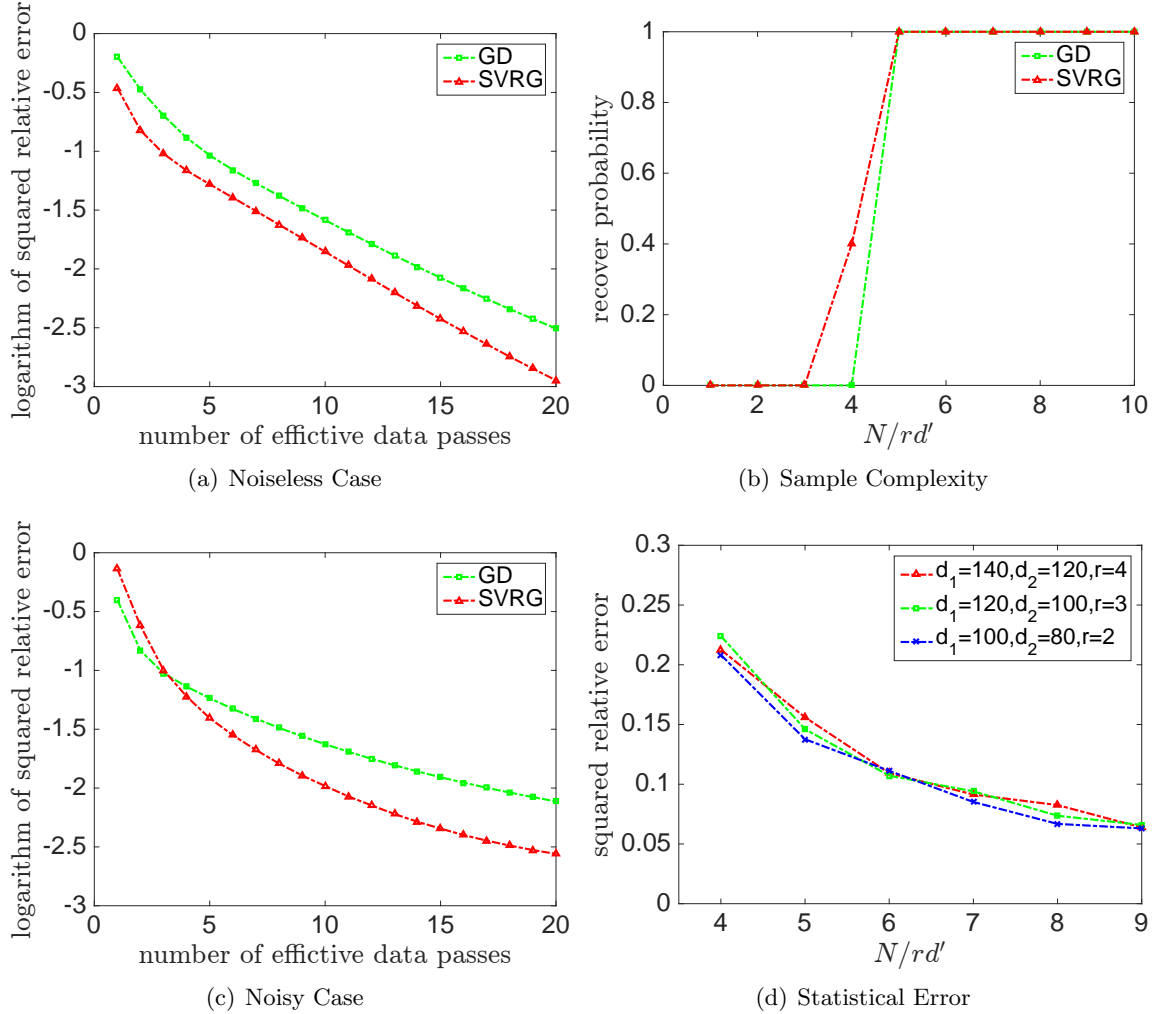

Figure 1: Numerical results for matrix sensing. (a) and (c) Convergence rates for matrix sensing in the noiseless and noisy case, respectively: logarithm of $\|\widehat{\mathbf{X}} - \mathbf{X}^*\|_F^2/\|\mathbf{X}^*\|_F^2$ versus number of effective data passes. They demonstrate the linear convergence rate and the superiority of our method; (b) Empirical probability of exact recovery versus $N/(rd')$, which confirms the optimal sample complexity in the noiseless case that $N = O(rd')$; (d) Statistical error: $\|\widehat{\mathbf{X}} - \mathbf{X}^*\|_F^2/\|\mathbf{X}^*\|_F^2$ versus $N/(rd')$, which matches the statistical error of our theory.

For the results of convergence rate, we compute the logarithm of the squared relative error $\|\widehat{\mathbf{X}} - \mathbf{X}^*\|_F^2/\|\mathbf{X}^*\|_F^2$, which are displayed in Figure 3(a). The results not only confirm the linear rate of convergence of our algorithm, but also demonstrate the effectiveness of our method after the same number of effective data passes. Besides, since we get results with similar patterns for other

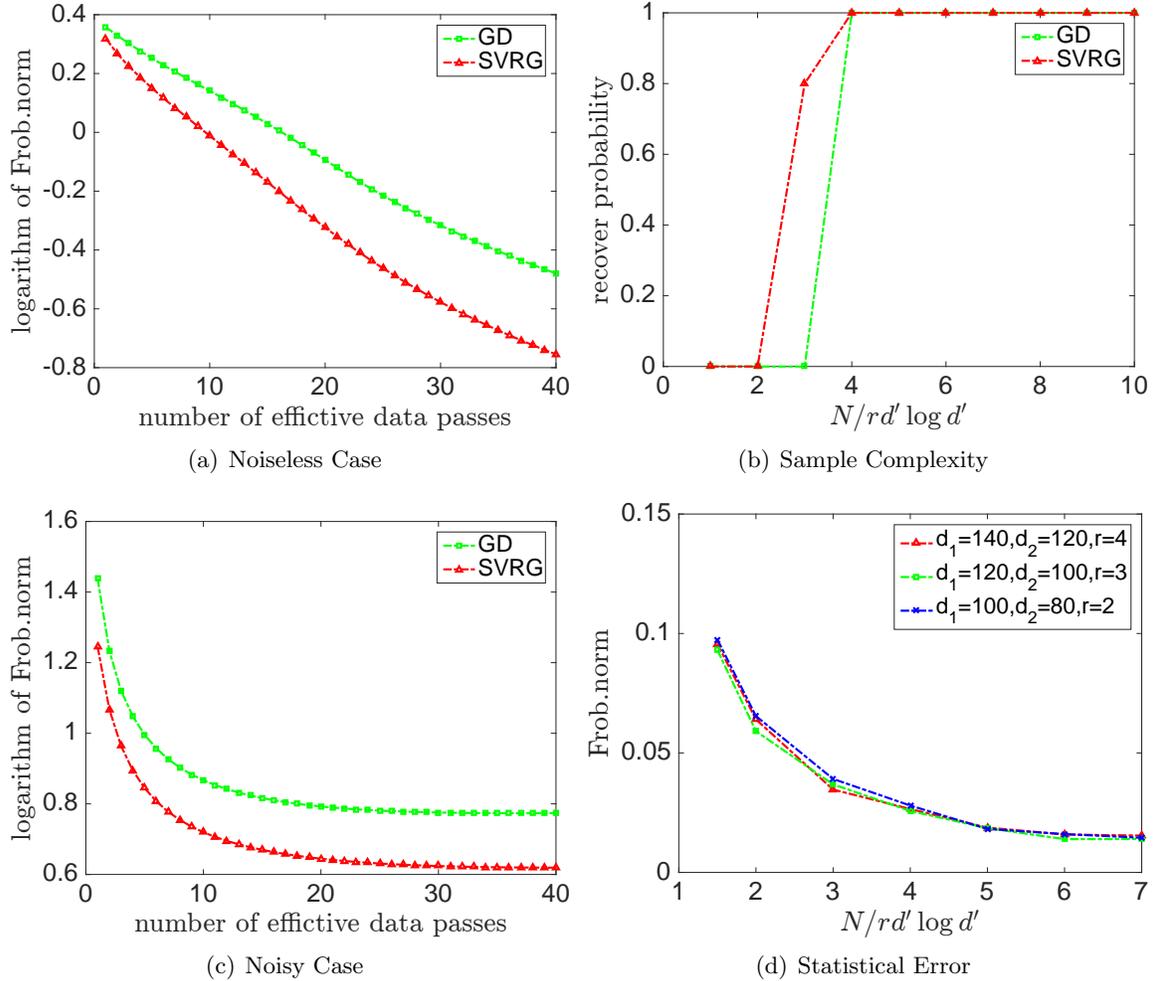

Figure 2: Numerical results for matrix completion. (a) and (c) Rate of convergence for matrix completion in the noiseless and noisy case, respectively: logarithm of mean squared error $\|\widehat{\mathbf{X}} - \mathbf{X}^*\|_F^2/(d_1 d_2)$ versus number of effective data passes, which demonstrates the effectiveness of our method; (b) Empirical probability of exact recovery versus $N/(rd')$, which confirms the optimal sample complexity in noiseless case that $N = O(rd' \log d')$; (d) Statistical error for matrix completion: mean squared error in Frobenius norm $\|\widehat{\mathbf{X}} - \mathbf{X}^*\|_F^2/(d_1 d_2)$ versus rescaled sample size $N/(rd' \log d')$, which is consistent with our theory.

settings, we leave them out for simplicity. For the results of statistical error, Figure 3(b) illustrates that with the same percentage of observations, the squared relative error decreases as the ratio $r/d$ decreases, which is consistent with the statistical error $O(rd'/|\Omega|)$.

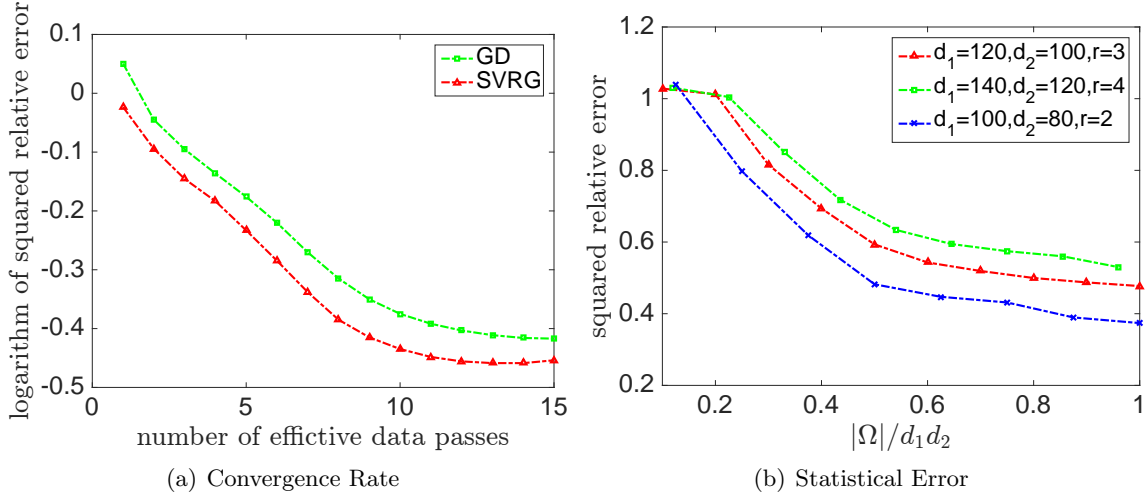

(a) Convergence Rate  (b) Statistical Error

Figure 3: Numerical results for one-bit matrix completion. (a) Convergence rates for one-bit matrix completion: logarithm of squared relative error $\|\widehat{\mathbf{X}} - \mathbf{X}^*\|_F^2 / \|\mathbf{X}^*\|_F^2$ versus number of effective data passes. It illustrates the superiority of our method after the same number of effective data passes; (b) Statistical error for one-bit matrix completion: squared relative error $\|\widehat{\mathbf{X}} - \mathbf{X}^*\|_F^2 / \|\mathbf{X}^*\|_F^2$ versus $|\Omega|/(d_1 d_2)$, which verifies the statistical rate.

# 7 Conclusions

We proposed a unified stochastic variance-reduced gradient descent framework for low-rank matrix recovery that integrates both optimization-theoretic and statistical analyses. We established the linear convergence rate of our proposed algorithm under mild restricted strong convexity and smoothness conditions. With an appropriate initialization algorithm, we proved that our algorithm enjoys improved computational complexity compared with existing approaches. We further illustrated the advantages of our unified framework through three specific examples: matrix sensing, matrix completion and one-bit matrix completion. Extensive experiments on synthetic data verified the strength of our algorithm and theory.

# A Proofs of Specific Models

## A.1 Proofs of Corollary 4.12

In order to prove the theoretical guarantees for matrix sensing, we only need to verify the restricted strong convexity and smoothness conditions for sample loss function $\mathcal{L}_N$, the restricted strong smoothness condition for each component function $\mathcal{L}_{\mathcal{S}_i}$ and the upper bound of $\|\nabla \mathcal{L}_N(\mathbf{X}^*)\|_2$.

First, we briefly introduce the definition of $\mathbf{\Sigma}$-ensemble which has been used in (Negahban and Wainwright, 2011) to verify the similar property of random sensing matrix $\mathbf{A}_i$ with dependent elements. Let $\text{vec}(\mathbf{A}_i) \in \mathbb{R}^{d_1 d_2}$ be the vectorization of sensing matrix $\mathbf{A}_i$. If $\text{vec}(\mathbf{A}_i) \sim N(0, \mathbf{\Sigma})$, we say that the sensing matrix $\mathbf{A}_i$ is sampled from $\mathbf{\Sigma}$-ensemble. In addition, we define $\pi^2(\mathbf{\Sigma}) = \sup_{\|\mathbf{u}\|_2=1, \|\mathbf{v}\|_2=1} \text{Var}(\mathbf{u}^\top \mathbf{A} \mathbf{v})$. Specifically, in classical matrix sensing, we have $\mathbf{\Sigma} = \mathbf{I}$ and $\pi(\mathbf{I}) = 1$.

Recall that we have linear measurement operators $\mathcal{A}_N(\mathbf{X}) = (\langle \mathbf{A}_1, \mathbf{X} \rangle, \langle \mathbf{A}_2, \mathbf{X} \rangle, \ldots, \langle \mathbf{A}_N, \mathbf{X} \rangle)^\top$, and $\mathcal{A}_{\mathcal{S}_i}(\mathbf{X}) = (\langle \mathbf{A}_{i_1}, \mathbf{X}\rangle, \langle \mathbf{A}_{i_2}, \mathbf{X}\rangle, \ldots, \langle \mathbf{A}_{i_b}, \mathbf{X}\rangle)^\top$ for $i = 1, \ldots, n$. In order to prove the restricted strongly convex and smooth conditions of our objective function, we need to ultilize the following lemma, which has been used in Agarwal et al. (2010); Negahban and Wainwright (2011).

**Lemma A.1.** Suppose each sensing matrix $\mathbf{A}_i$ of the linear measurement operator $\mathcal{A}_M$ is sampled from $\boldsymbol{\Sigma}$-ensemble, where $M$ is the number of sensing matrices. Then there exist constants $C_0$ and $C_1$, such that the following inequalities hold for all $\boldsymbol{\Delta} \in \mathbb{R}^{d_1 \times d_2}$ with probability at least $1 - \exp(-C_0 M)$

$$\frac{\|\mathcal{A}(\boldsymbol{\Delta})\|_2^2}{M} \geq \frac{1}{2}\|\sqrt{\boldsymbol{\Sigma}}\mathrm{vec}(\boldsymbol{\Delta})\|_2^2 - C_1 \pi^2(\boldsymbol{\Sigma})\frac{d'}{M}\|\boldsymbol{\Delta}\|_*^2, \tag{A.1}$$

$$\frac{\|\mathcal{A}(\boldsymbol{\Delta})\|_2^2}{M} \leq \frac{1}{2}\|\sqrt{\boldsymbol{\Sigma}}\mathrm{vec}(\boldsymbol{\Delta})\|_2^2 + C_1 \pi^2(\boldsymbol{\Sigma})\frac{d'}{M}\|\boldsymbol{\Delta}\|_*^2. \tag{A.2}$$

In order to bound the gradient of sample loss function $\nabla \mathcal{L}_M(\mathbf{X}^*)$ with respect to $M$ observations, we need to ultilize the following lemma, which has been used in Negahban and Wainwright (2011).

**Lemma A.2.** Suppose each sensing matrix $\mathbf{A}_i$ of the linear measurement operator $\mathcal{A}_M$ is sampled from $\boldsymbol{\Sigma}$-ensemble, where $M$ is the number of sensing matrices. Furthermore, suppose the noise vector $\boldsymbol{\epsilon}$ satisfies that $\|\boldsymbol{\epsilon}\|_2 \leq 2\nu\sqrt{M}$. Then we have the following inequality

$$\left\|\frac{1}{M}\sum_{i=1}^M \epsilon_i \mathbf{A}_i\right\|_2 \leq C\nu\sqrt{\frac{d'}{M}},$$

holds with probability at least $1 - C_1\exp(-C_2 d')$, where $C, C_1$ and $C_2$ are universal constants.

Note that Lemma A.2 requires the noise vector $\boldsymbol{\epsilon}$ satisfies $\|\boldsymbol{\epsilon}\|_2 \leq 2\nu\sqrt{M}$ for some constant $\nu$. For any bounded noise vector, this condition obviously holds. And if the noise vector follows sub-Gaussian distribution with parameter $\nu$, it has been proved in Vershynin (2010) that this condition holds with high probability.

*Proof of Corollary 4.12.* First, we prove the restricted strong convexity condition for sample loss function $\mathcal{L}_N$. First, we have

$$\mathcal{L}_N(\mathbf{X}) = \frac{1}{2N}\sum_{i=1}^N \left(\langle \mathbf{A}_i, \mathbf{X}\rangle - y_i\right)^2 = \frac{1}{2N}\sum_{i=1}^N \left(\langle \mathbf{A}_i, \mathbf{X} - \mathbf{X}^*\rangle - \epsilon_i\right)^2.$$

Consider two rank-$r$ matrices $\mathbf{X}, \mathbf{Y} \in \mathbb{R}^{d_1 \times d_2}$. Let $\boldsymbol{\Delta} = \mathbf{Y} - \mathbf{X}$, then we have the following equality

$$\mathcal{L}_N(\mathbf{Y}) - \mathcal{L}_N(\mathbf{X}) - \langle \nabla \mathcal{L}_N(\mathbf{X}), \boldsymbol{\Delta}\rangle$$
$$= \frac{1}{2N}\sum_{i=1}^N \left(\langle \mathbf{A}_i, \mathbf{Y} - \mathbf{X}^*\rangle^2 - \langle \mathbf{A}_i, \mathbf{X} - \mathbf{X}^*\rangle^2 - 2\langle \mathbf{A}_i, \mathbf{X} - \mathbf{X}^*\rangle\langle \mathbf{A}_i, \boldsymbol{\Delta}\rangle\right)$$
$$= \frac{\|\mathcal{A}(\boldsymbol{\Delta})\|_2^2}{2N}. \tag{A.3}$$

Therefore, according to (A.3), in order to establish the restricted strongly convex and smooth conditions for $\mathcal{L}_N$, we need to bound the term $\|\mathcal{A}(\boldsymbol{\Delta})\|_2^2/N$. According to (A.1) in Lemma A.1, we get

$$\frac{\|\mathcal{A}(\boldsymbol{\Delta})\|_2^2}{N} \geq \frac{1}{2}\|\sqrt{\boldsymbol{\Sigma}}\mathrm{vec}(\boldsymbol{\Delta})\|_2^2 - C_1\pi^2(\boldsymbol{\Sigma})\frac{d'}{N}\|\boldsymbol{\Delta}\|_*^2.$$

Furthermore, note that $\boldsymbol{\Delta} = \mathbf{Y} - \mathbf{X}$ has rank at most $2r$. Thus, we conclude that $\|\boldsymbol{\Delta}\|_* \leq \sqrt{2r}\|\boldsymbol{\Delta}\|_F$, which further implies

$$\frac{\|\mathcal{A}(\boldsymbol{\Delta})\|_2^2}{N} \geq \left\{\frac{\lambda_{\min}(\boldsymbol{\Sigma})}{2} - 2C_1 r \pi^2(\boldsymbol{\Sigma})\frac{d'}{N}\right\}\|\boldsymbol{\Delta}\|_F^2.$$

Therefore, as long as $N \geq C_3 \pi^2(\boldsymbol{\Sigma}) r d'/\lambda_{\min}(\boldsymbol{\Sigma})$ for some sufficiently large constant $C_3$, we get

$$\frac{\|\mathcal{A}(\boldsymbol{\Delta})\|_2^2}{N} \geq \frac{4\lambda_{\min}(\boldsymbol{\Sigma})}{9}\|\boldsymbol{\Delta}\|_F^2.$$

Since $\boldsymbol{\Sigma} = \mathbf{I}$ in matrix sensing, we obtain the restricted strongly convex parameter $\mu = 4/9$.

Second, we prove the restricted strong smoothness condition for $\mathcal{L}_N$ using (A.2) in Lemma A.1. Similar to the proof of the restricted strong convexity condition, we get

$$\frac{\|\mathcal{A}(\boldsymbol{\Delta})\|_2^2}{N} \leq \frac{5\lambda_{\max}(\boldsymbol{\Sigma})}{9}\|\boldsymbol{\Delta}\|_F^2,$$

as long as $N \geq C_3 \pi^2(\boldsymbol{\Sigma}) r d'/\lambda_{\min}(\boldsymbol{\Sigma})$ for some sufficiently large constant $C_3$. Therefore, because $\boldsymbol{\Sigma} = \mathbf{I}$ in matrix sensing, we accordingly obtain $L = 5/9$.

Next, we prove each component loss function $\mathcal{L}_{\mathcal{S}_i}$ is restricted strongly smooth, for $i = 1, \ldots, n$. Recall that we have

$$\mathcal{L}_{\mathcal{S}_i}(\mathbf{X}) = \frac{1}{2b}\|\mathbf{y}_{\mathcal{S}_i} - \mathcal{A}_{\mathcal{S}_i}(\mathbf{U}\mathbf{V}^\top)\|_2^2 = \frac{1}{2b}\sum_{j \in \Omega_{\mathcal{S}_i}}\left(\langle\mathbf{A}_j, \mathbf{X} - \mathbf{X}^*\rangle - \epsilon_j\right)^2.$$

Thus, for each component loss function $\mathcal{L}_{\mathcal{S}_i}$, where $i = 1, \ldots, n$, we have

$$\mathcal{L}_{\mathcal{S}_i}(\mathbf{Y}) - \mathcal{L}_{\mathcal{S}_i}(\mathbf{X}) - \langle\nabla\mathcal{L}_{\mathcal{S}_i}(\mathbf{X}), \boldsymbol{\Delta}\rangle = \frac{\|\mathcal{A}_{\mathcal{S}_i}(\boldsymbol{\Delta})\|_2^2}{2b}. \tag{A.4}$$

Following the same steps as in the proof of restricted strong smoothness condition for $\mathcal{L}_N$, for each component function $\mathcal{L}_{\mathcal{S}_i}$, we get

$$\frac{\|\mathcal{A}_{\mathcal{S}_i}(\boldsymbol{\Delta})\|_2^2}{b} \leq C_5 \lambda_{\max}(\boldsymbol{\Sigma})\|\boldsymbol{\Delta}\|_F^2,$$

if $b \geq C_4 \pi^2(\boldsymbol{\Sigma}) r d'/\lambda_{\min}(\boldsymbol{\Sigma})$ for some sufficiently large constant $\mathcal{C}_4$. Therefore we have $L' = C_5$ since $\boldsymbol{\Sigma} = \mathbf{I}$.

Finally, we bound the statistical error term $\|\nabla\mathcal{L}_N(\mathbf{X}^*)\|_2^2$. According to the definition of $\mathcal{L}_N$, we have

$$\nabla\mathcal{L}_N(\mathbf{X}^*) = \frac{1}{N}\sum_{i=1}^N \epsilon_i \mathbf{A}_i.$$

Based on Lemma A.2, we have the following inequality holds with probability at least $1 - C'_1\exp(-C'_2 d')$

$$\left\|\frac{1}{N}\sum_{i=1}^N \epsilon_i \mathbf{A}_i\right\|_2 \leq C\nu\sqrt{\frac{d'}{N}},$$

which implies that

$$\|\nabla\mathcal{L}_N(\mathbf{X}^*)\|_2^2 \leq C^2 \nu^2 \frac{d'}{N}.$$

$\square$

## A.2 Proofs of Corollary 4.14

In order to prove the theoretical guarantees for matrix completion we only need to verify the restricted strong convexity and smoothness conditions for sample loss function $\mathcal{L}_\Omega$, the restricted strong smoothness condition for each component function $\mathcal{L}_{\Omega_{\mathcal{S}_i}}$ and the upper bound of $\|\nabla\mathcal{L}_\Omega(\mathbf{X}^*)\|_2$.

We first briefly introduce the definition of spikiness constraint (Negahban and Wainwright, 2012; Gunasekar et al., 2014), which is essential for us to provide the restricted strong convexity and smoothness conditions. The spikiness ratio is defined as $\alpha_{\rm sp}(\mathbf{X}) = (\sqrt{d_1 d_2}\|\mathbf{X}\|_\infty)/\|\mathbf{X}\|_F$. This ratio can be used to avoid the overly spiky matrices, which are impossible to recover unless we observe all elements of the unknown matrix (Gross, 2011). Thus we impose following spikiness constraint for low-rank matrices in matrix completion $\|\mathbf{X}\|_\infty \leq \alpha$, where $\alpha = \alpha_{\rm sp}(\mathbf{X})\|\mathbf{X}\|_F/\sqrt{d_1 d_2}$. Note that this condition can be satisfied by using projection in our algorithm. Note that some existing work (Candès and Recht, 2009) assumes incoherence conditions to preclude such overly spiky matrices. However, incoherence conditions are more restrictive than the spikiness constraint (Negahban and Wainwright, 2012).

To establish the restricted strong convexity and smoothness conditions for $\mathcal{L}_\Omega$, and the restricted strong smoothness condition for $\mathcal{L}_{\Omega_{\mathcal{S}_i}}$ under the spikiness constraint, we need to ultilize the following lemma, which used in Negahban and Wainwright (2012).

**Lemma A.3.** Suppose the number of observations $M$ satisfying $M > c_1 d' \log d'$. Furthermore, if for all $\mathbf{\Delta} \in \mathbb{R}^{d_1 \times d_2}$, we have

$$\alpha_{\rm sp}(\mathbf{\Delta})\frac{\|\mathbf{\Delta}\|_*}{\|\mathbf{\Delta}\|_F} \leq \frac{1}{c_2}\sqrt{M/d'\log d'}, \tag{A.5}$$

then the following inequality holds with probability at least $1 - c_3\exp(-c_4 d' \log d')$

$$\left|\frac{\|\mathcal{A}(\mathbf{\Delta})\|_2}{\sqrt{M}} - \frac{\|\mathbf{\Delta}\|_F}{\sqrt{d_1 d_2}}\right| \leq \frac{1}{10}\frac{\|\mathbf{\Delta}\|_F}{\sqrt{d_1 d_2}}\left(1 + \frac{c_5\alpha_{\rm sp}(\mathbf{\Delta})}{\sqrt{M}}\right),$$

where $c_1, c_2, c_3, c_4, c_5$ are universal constants.

Moreover, in order to upper bound of the gradient $\nabla\mathcal{L}_\Omega$ at $\mathbf{X}^*$, we need to use the following lemma.

**Lemma A.4.** (Negahban and Wainwright, 2012) Suppose $\mathbf{A}_i$ is uniformly distributed over $\mathcal{X}$. In addition, each noise $\epsilon_i$ follows i.i.d. normal distribution with variance $\nu^2$. Then the following inequality holds with probability at least $1 - c_1/d'$

$$\left\|\frac{1}{M}\sum_{i=1}^{M}\epsilon_i\mathbf{A}_i\right\|_2 \leq c_2\nu\sqrt{\frac{d'\log d'}{d_1 d_2 M}},$$

where $M$ is the number of observations, and $c_1, c_2$ are universal constants.

*Proof of Corollary 4.14.* Let $|\Omega| = N$, $|\Omega_{\mathcal{S}_i}| = b$. For any $(j,k) \in \Omega$, we denote $\mathbf{A}_{jk} = \mathbf{e}_j\mathbf{e}_k^\top$, where $\mathbf{e}_j, \mathbf{e}_k$ are unit vectors with dimensionality $d_1$ and $d_2$ respectively. Similarly, for any $(j,k) \in \Omega_{\mathcal{S}_i}$, we let $\mathbf{A}_{jk}^i = \mathbf{e}_j^i\mathbf{e}_k^{i\top}$. Thus, we can rewrite the sample loss function as follows (here for simplicity, we use $\mathcal{L}_N$ and $\mathcal{L}_{\mathcal{S}_i}$ to denote $\mathcal{L}_\Omega$ and $\mathcal{L}_{\Omega_{\mathcal{S}_i}}$ respectively)

$$\mathcal{L}_N(\mathbf{X}) := \frac{1}{2p}\sum_{(j,k)\in\Omega}\left(\langle\mathbf{A}_{jk},\mathbf{X}\rangle - Y_{jk}\right)^2,$$

where $p = N/(d_1 d_2)$. In addition, we can rewrite each component loss function as follows

$$\mathcal{L}_{\mathcal{S}_i}(\mathbf{X}) := \frac{1}{2p'} \sum_{(j,k) \in \Omega_{\mathcal{S}_i}} \left( \langle \mathbf{A}_{jk}^i, \mathbf{X} \rangle - Y_{jk} \right)^2,$$

where $p' = b/(d_1 d_2)$. For simplicity we let $\mathcal{A}$ and $\mathcal{A}_{\mathcal{S}_i}$ be the corresponding transformation operator with respect to $\mathcal{L}_N$ and $\mathcal{L}_{\mathcal{S}_i}$, respectively. First, we prove the restricted strong convexity and smoothness conditions for $\mathcal{L}_N$. Recall that we have the spikiness constraint set $\mathcal{C}(\alpha) = \{ \mathbf{X} \in \mathbb{R}^{d_1 \times d_2} : \|\mathbf{X}\|_\infty \leq \alpha \}$. Consider any two rank-$r$ matrices $\mathbf{X}, \mathbf{Y}$, which satisfy the spikiness constraint that $\mathbf{X}, \mathbf{Y} \in \mathcal{C}(\alpha)$. In the following discussion, denote $\mathbf{\Delta} = \mathbf{Y} - \mathbf{X}$.

**Case 1:** If condition (A.5) is violated. Then we obtain

$$\|\mathbf{\Delta}\|_F^2 \leq C_0 \left( \sqrt{d_1 d_2} \|\mathbf{\Delta}\|_\infty \right) \cdot \|\mathbf{\Delta}\|_* \sqrt{\frac{d' \log d'}{N}}$$

$$\leq 2 C_0 \alpha \sqrt{d_1 d_2} \|\mathbf{\Delta}\|_* \sqrt{\frac{d' \log d'}{N}},$$

where the last inequality is due to the fact that $\|\mathbf{\Delta}\|_\infty = \|\mathbf{X} - \mathbf{Y}\|_\infty \leq 2\alpha$. Furthermore, we get

$$\|\mathbf{\Delta}\|_F^2 \leq 2 C_0 \sqrt{2r} \alpha \sqrt{d_1 d_2} \|\mathbf{\Delta}\|_F \sqrt{\frac{d' \log d'}{N}},$$

where the inequality holds because $\text{rank}(\mathbf{\Delta}) \leq 2r$, which implies the following bound

$$\frac{1}{d_1 d_2} \|\mathbf{\Delta}\|_F^2 \leq C \alpha^2 \frac{r d' \log d'}{N}. \tag{A.6}$$

**Case 2:** If condition (A.5) is satisfied. We first establish the restricted strongly convex condition for $\mathcal{L}_N$. In particular, we have

$$\mathcal{L}_N(\mathbf{Y}) - \mathcal{L}_N(\mathbf{X}) - \langle \nabla \mathcal{L}_N(\mathbf{X}), \mathbf{\Delta} \rangle$$
$$= \frac{1}{2p} \sum_{(j,k) \in \Omega} \left( \langle \mathbf{A}_{jk}, \mathbf{Y} - \mathbf{X}^* \rangle^2 + \langle \mathbf{A}_{jk}, \mathbf{X} - \mathbf{X}^* \rangle^2 - 2 \langle \mathbf{A}_{jk}, \mathbf{X} - \mathbf{X}^* \rangle \langle \mathbf{A}_{jk}, \mathbf{\Delta} \rangle \right)$$
$$= \frac{\|\mathcal{A}(\mathbf{\Delta})\|_2^2}{2p}, \tag{A.7}$$

Thus, as long as $c_5 \alpha_{\text{sp}}(\mathbf{\Delta}) \geq \sqrt{N}$, by the definition of spikiness ration $\alpha_{\text{sp}}(\mathbf{\Delta})$, we get

$$\frac{1}{d_1 d_2} \|\mathbf{\Delta}\|_F^2 \leq c' \alpha^2 \frac{1}{N}. \tag{A.8}$$

If $c_5 \alpha_{\text{sp}}(\mathbf{\Delta}) \leq \sqrt{N}$, according to Lemma A.3, we obtain

$$\frac{\|\mathcal{A}(\mathbf{\Delta})\|_2^2}{p} \geq \frac{8}{9} \|\mathbf{\Delta}\|_F^2,$$

which implies the restricted strong convexity parameter $\mu = 8/9$.

Next, for sample loss function $\mathcal{L}_N$, we establish the restricted strong smoothness condition by similar proof. According to (A.7) and Lemma A.3, as long as $c_5\alpha_{\text{sp}}(\Delta) \leq \sqrt{N}$, we have

$$\frac{\|\mathcal{A}(\Delta)\|_2^2}{p} \leq \frac{10}{9}\|\Delta\|_F^2,$$

which gives us the restricted strong smoothness parameter $L = 10/9$.

Similarly, we show the restricted strong smoothness condition for each component loss function $\mathcal{L}_{\mathcal{S}_i}$, where $i = 1, \ldots, n$. Since we have

$$\mathcal{L}_{\mathcal{S}_i}(\mathbf{Y}) - \mathcal{L}_{\mathcal{S}_i}(\mathbf{X}) - \langle \nabla \mathcal{L}_{\mathcal{S}_i}(\mathbf{X}), \Delta \rangle = \frac{\|\mathcal{A}_{\mathcal{S}_i}(\Delta)\|_2^2}{2p'},$$

thus according to Lemma A.3, as long as $c_5\alpha_{\text{sp}}(\Delta)/\sqrt{b} \leq c_6$, we have

$$\frac{\|\mathcal{A}_{\mathcal{S}_i}(\Delta)\|_2^2}{p'} \leq c_7\|\Delta\|_F^2,$$

which implies that $L' = c_7$. Otherwise, it is sufficient to ensure $\alpha = O(1/\sqrt{n})$.

Finally, for the statistical error term $\|\nabla \mathcal{L}_N(\mathbf{X}^*)\|_2^2$, according to the definition of $\mathcal{L}_N$, we have

$$\nabla \mathcal{L}_N(\mathbf{X}^*) = \frac{1}{p} \sum_{(j,k) \in \Omega} \epsilon_{jk} \mathbf{A}_{jk}.$$

Remember that each elements of the noise matrix follows i.i.d. Gaussian distribution with variance $\nu^2$. Therefore, according to Lemma A.4, we obtain

$$\left\| \frac{1}{p} \sum_{(j,k) \in \Omega} \epsilon_{jk} \mathbf{A}_{jk} \right\|_2 \leq C\nu\sqrt{d_1 d_2}\sqrt{\frac{d' \log d'}{N}},$$

holds with probability at least $1 - C'/d'$, which implies that

$$\|\nabla \mathcal{L}_N(\mathbf{X}^*)\|_2^2 \leq C^2\nu^2 d_1 d_2 \frac{d' \log d'}{N}, \tag{A.9}$$

holds with probability at least $1 - C'/d'$. Combining error bounds (A.6), (A.8) and (A.9), we conclude the following statistical error

$$C_1 \max\{\alpha^2, \nu^2\} \frac{d_1 d_2 d' \log d'}{|\Omega|}.$$

$\square$

## A.3 Proofs of Corollary 4.16

In order to prove the theoretical guarantees for one-bit matrix completion, we only need to verify the restricted strong convexity and smoothness conditions for the sample loss function $\mathcal{L}_\Omega$, the restricted strong smoothness condition for each component function $\mathcal{L}_{\Omega_{\mathcal{S}_i}}$, and the upper bound of $\|\nabla \mathcal{L}_\Omega(\mathbf{X}^*)\|_2$. In order to establish the statistical error bound, we need to ultilize the following lemma, which has been used in Negahban and Wainwright (2012).

**Lemma A.5.** Suppose observation index set $\Omega \subseteq [d_1] \times [d_2]$ is sampled under uniform model, then there exist universal constants $C, C'$ such that the following inequality holds with probability at least $1 - C'/d'$.

$$\left\| \frac{1}{|\Omega|} \sum_{j,k \in \Omega} \sqrt{d_1 d_2} \mathbf{e}_j \mathbf{e}_k^\top \right\|_2 \leq C \sqrt{\frac{d' \log d'}{|\Omega|}}.$$

*Proof of Corollary 4.16.* Let $|\Omega| = N$, $|\Omega_{\mathcal{S}_i}| = b$. For any $(j,k) \in \Omega$, we denote $\mathbf{A}_{jk} = \mathbf{e}_j \mathbf{e}_k^\top$, where $\mathbf{e}_i, \mathbf{e}_j$ are unit vectors with $d_1$ and $d_2$ dimensions. Similarly, for any $(j,k) \in \Omega_{\mathcal{S}_i}$, we let $\mathbf{A}_{jk}^i = \mathbf{e}_j^i \mathbf{e}_k^{i\top}$. Note that for simplicity we use $\mathcal{A}$ and $\mathcal{A}_{\mathcal{S}_i}$ to denote the corresponding transformation operator with respect to $\mathcal{L}_N$ and $\mathcal{L}_{\mathcal{S}_i}$, respectively. We can rewrite the sample loss function as follows (here for simplicity, we use $\mathcal{L}_N$ and $\mathcal{L}_{\mathcal{S}_i}$ to denote $\mathcal{L}_\Omega$ and $\mathcal{L}_{\Omega_{\mathcal{S}_i}}$ respectively)

$$\mathcal{L}_N(\mathbf{X}) := -\frac{1}{p} \sum_{(j,k) \in \Omega} \left\{ \mathbb{1}_{(Y_{jk}=1)} \log \left( f(\langle \mathbf{A}_{jk}, \mathbf{X} \rangle) \right) + \mathbb{1}_{(Y_{jk}=-1)} \log \left( 1 - f(\langle \mathbf{A}_{jk}, \mathbf{X} \rangle) \right) \right\},$$

where $p = N/d_1 d_2$. Therefore, we have each component loss function $\mathcal{L}_{\mathcal{S}_i}$ as

$$\mathcal{L}_{\mathcal{S}_i}(\mathbf{X}) := -\frac{1}{p'} \sum_{(j,k) \in \Omega_{\mathcal{S}_i}} \left\{ \mathbb{1}_{(Y_{jk}=1)} \log \left( f(\langle \mathbf{A}_{jk}^i, \mathbf{X} \rangle) \right) + \mathbb{1}_{(Y_{jk}=-1)} \log \left( 1 - f(\langle \mathbf{A}_{jk}^i, \mathbf{X} \rangle) \right) \right\},$$

where $p' = b/d_1 d_2$. Therefore, we get

$$\nabla \mathcal{L}_N(\mathbf{X}) = \frac{1}{p} \sum_{(j,k) \in \Omega} \left( -\frac{f'(\langle \mathbf{A}_{jk}, \mathbf{X} \rangle)}{f(\langle \mathbf{A}_{jk}, \mathbf{X} \rangle)} \mathbb{1}_{(Y_{jk}=1)} + \frac{f'(\langle \mathbf{A}_{jk}, \mathbf{X} \rangle)}{1 - f(\langle \mathbf{A}_{jk}, \mathbf{X} \rangle)} \mathbb{1}_{(Y_{jk}=-1)} \right) \mathbf{A}_{jk}. \quad (A.10)$$

Furthermore, we obtain

$$\nabla^2 \mathcal{L}_N(\mathbf{X}) = \frac{1}{p} \sum_{(j,k) \in \Omega} B_{jk}(\mathbf{X}) \operatorname{vec}(\mathbf{A}_{jk}) \operatorname{vec}(\mathbf{A}_{jk})^\top, \quad (A.11)$$

where we have

$$B_{jk}(\mathbf{X}) = \left[ \left( \frac{f'^2(\langle \mathbf{A}_{jk}, \mathbf{X} \rangle)}{f^2(\langle \mathbf{A}_{jk}, \mathbf{X} \rangle)} - \frac{f''(\langle \mathbf{A}_{jk}, \mathbf{X} \rangle)}{f(\langle \mathbf{A}_{jk}, \mathbf{X} \rangle)} \right) \mathbb{1}_{(Y_{jk}=1)} + \left( \frac{f''(\langle \mathbf{A}_{jk}, \mathbf{X} \rangle)}{1 - f(\langle \mathbf{A}_{jk}, \mathbf{X} \rangle)} - \frac{f'^2(\langle \mathbf{A}_{jk}, \mathbf{X} \rangle)}{(1 - f(\langle \mathbf{A}_{jk}, \mathbf{X} \rangle))^2} \right) \mathbb{1}_{(Y_{jk}=-1)} \right].$$

First, we establish the strong convexity and smoothness conditions for $\mathcal{L}_N$. For any $\mathbf{X}, \mathbf{M} \in \mathbb{R}^{d_1 \times d_2}$, let $\mathbf{W} = \mathbf{M} + a(\mathbf{X} - \mathbf{M})$ for $a \in [0, 1]$, $\mathbf{x} = \operatorname{vec}(\mathbf{X})$ and $\mathbf{m} = \operatorname{vec}(\mathbf{M})$. According to the mean value theorem, we get

$$\mathcal{L}_N(\mathbf{X}) = \mathcal{L}_N(\mathbf{M}) + \langle \nabla \mathcal{L}_N(\mathbf{M}), \mathbf{X} - \mathbf{M} \rangle + \frac{1}{2}(\mathbf{x} - \mathbf{m})^\top \nabla^2 \mathcal{L}_N(\mathbf{W})(\mathbf{x} - \mathbf{m}),$$

Moreover, according to (A.11), we further obtain

$$(\mathbf{x} - \mathbf{m})^\top \nabla^2 \mathcal{L}_N(\mathbf{W})(\mathbf{x} - \mathbf{m}) = \frac{1}{p} \sum_{(j,k) \in \Omega} B_{jk}(\mathbf{W}) \langle \operatorname{vec}(\mathbf{A}_{jk})^\top (\mathbf{x} - \mathbf{m}), \operatorname{vec}(\mathbf{A}_{jk})^\top (\mathbf{x} - \mathbf{m}) \rangle$$

$$= \frac{1}{p} \sum_{(j,k) \in \Omega} B_{jk}(\mathbf{W}) \langle \mathbf{A}_{jk}, \boldsymbol{\Delta} \rangle^2,$$

where $\boldsymbol{\Delta} = \mathbf{X} - \mathbf{M}$. Thus, according to the definition of $\mu_\alpha$ in (4.5), we obtain

$$\frac{1}{p} \sum_{(j,k)\in\Omega} B_{jk}(\mathbf{W})\langle \mathbf{A}_{jk}, \boldsymbol{\Delta}\rangle^2 \geq \mu_\alpha \frac{\|\mathcal{A}(\boldsymbol{\Delta})\|_2^2}{p},$$

Therefore, following the same steps in the proof of matrix completion, we have

$$\mu_\alpha \frac{\|\mathcal{A}(\boldsymbol{\Delta})\|_2^2}{p} \geq C_1 \mu_\alpha \|\boldsymbol{\Delta}\|_F^2,$$

which implies

$$\mathcal{L}_N(\mathbf{X}) \geq \mathcal{L}_N(\mathbf{M}) + \langle \nabla \mathcal{L}_N(\mathbf{M}), \mathbf{X} - \mathbf{M}\rangle + \frac{1}{2} C_1 \mu_\alpha \|\boldsymbol{\Delta}\|_F^2.$$

Therefore it gives us the restricted strong convexity parameter $\mu = C_1 \mu_\alpha$. On the other hand, according to the definition of $L_\alpha$ in (4.6), we obtain

$$\frac{1}{p} \sum_{(j,k)\in\Omega} B_{jk}(\mathbf{W})\langle \mathbf{A}_{jk}, \boldsymbol{\Delta}\rangle^2 \leq L_\alpha \frac{\|\mathcal{A}(\boldsymbol{\Delta})\|_2^2}{p},$$

Therefore, following the same steps in the proof of the restricted strong smoothness condition in matrix completion, we get

$$\mathcal{L}_N(\mathbf{X}) \leq \mathcal{L}_N(\mathbf{M}) + \langle \nabla \mathcal{L}_N(\mathbf{M}), \mathbf{X} - \mathbf{M}\rangle + \frac{1}{2} C_2 L_\alpha \|\boldsymbol{\Delta}\|_F^2,$$

which implies the restricted strong smoothness parameter $L = C_2 L_\alpha$. By the similar procedure, for each component function, we can derive that

$$\mathcal{L}_{\mathcal{S}_i}(\mathbf{X}) \leq \mathcal{L}_{\mathcal{S}_i}(\mathbf{M}) + \langle \nabla \mathcal{L}_{\mathcal{S}_i}(\mathbf{M}), \mathbf{X} - \mathbf{M}\rangle + \frac{1}{2} C_3 L_\alpha \|\boldsymbol{\Delta}\|_F^2.$$

Finally, for statistical error term $\|\nabla \mathcal{L}_N(\mathbf{X}^*)\|_2^2$, according to (A.10), we get

$$\nabla \mathcal{L}_N(\mathbf{X}^*) = \frac{1}{p} \sum_{(j,k)\in\Omega} b_{jk} \mathbf{A}_{jk},$$

where we have

$$b_{jk} = -\frac{f'(\langle \mathbf{A}_{jk}, \mathbf{X}^*\rangle)}{f(\langle \mathbf{A}_{jk}, \mathbf{X}^*\rangle)} \mathbb{1}_{(Y_{jk}=1)} + \frac{f'(\langle \mathbf{A}_{jk}, \mathbf{X}^*\rangle)}{1 - f(\langle \mathbf{A}_{jk}, \mathbf{X}^*\rangle)} \mathbb{1}_{(Y_{jk}=-1)}.$$

Therefore, according to the definition of $\gamma_\alpha$ in (4.7), we have

$$\|\nabla \mathcal{L}_N(\mathbf{X}^*)\|_2 = \frac{1}{p} \Big\| \sum_{(j,k)\in\Omega} b_{jk} \mathbf{A}_{jk} \Big\|_2 \leq \frac{1}{p} \gamma_\alpha \Big\| \sum_{(j,k)\in\Omega} \mathbf{A}_{jk} \Big\|_2,$$

Thus, by Lemma A.5, we obtain

$$\|\nabla \mathcal{L}_N(\mathbf{X}^*)\|_2 \leq C\gamma_\alpha \sqrt{\frac{d_1 d_2 d' \log d'}{|\Omega|}},$$

holds with probability at least $1 - C'/d'$, which is equivalent to

$$\|\nabla \mathcal{L}_N(\mathbf{X}^*)\|_2^2 \leq C^2 \gamma_\alpha^2 \frac{d' \log d'}{p}. \tag{A.12}$$

In addition, we also have the following bounds, which have been shown in proofs of matrix completion when condition (A.5) is not satisfied

$$\frac{1}{d_1 d_2} \|\boldsymbol{\Delta}\|_F^2 \leq \max\{C\alpha^2 \frac{1}{|\Omega|}, C'\alpha^2 \frac{rd' \log d'}{|\Omega|}\}. \tag{A.13}$$

Therefore, combining error bounds (A.13) and (A.12), we have the following statistical error

$$C \max\{\alpha^2, \gamma_\alpha^2\} \frac{d_1 d_2 d' \log d'}{|\Omega|}.$$

□

## B  Proofs of Technical Lemmas

In this section, we present the proofs of several technical lemmas. Before proceeding to the theoretical proof, we first introduce the following notations and definitions, which are essential for proving the following lemmas. For any $\mathbf{Z} \in \mathbb{R}^{(d_1+d_2) \times r}$, we denote $\mathbf{Z} = [\mathbf{U}; \mathbf{V}]$, where $\mathbf{U} \in \mathbb{R}^{d_1 \times r}$ and $\mathbf{V} \in \mathbb{R}^{d_2 \times r}$. Denote $\mathbf{X} = \mathbf{U}\mathbf{V}^\top$. Let $\mathbf{R} = \operatorname{argmin}_{\widetilde{\mathbf{R}} \in \mathbb{Q}_r} \|\mathbf{Z} - \mathbf{Z}^* \widetilde{\mathbf{R}}\|_F$ be the optimal rotation with respect to $\mathbf{Z}$, and $\mathbf{H} = \mathbf{Z} - \mathbf{Z}^*\mathbf{R} = [\mathbf{H}_U; \mathbf{H}_V]$, where $\mathbf{H}_U \in \mathbb{R}^{d_1 \times r}$, $\mathbf{H}_V \in \mathbb{R}^{d_2 \times r}$.

Moreover, let $\overline{\mathbf{U}}_1, \overline{\mathbf{U}}_2, \overline{\mathbf{U}}_3$ be the left singular matrices of $\mathbf{X}, \mathbf{U}, \mathbf{H}_U$, respectively. Define $\widetilde{\mathbf{U}}$ as the matrix spanned by the column of $\overline{\mathbf{U}}_1, \overline{\mathbf{U}}_2$ and $\overline{\mathbf{U}}_3$ such that

$$\operatorname{col}(\widetilde{\mathbf{U}}) = \operatorname{span}\{\overline{\mathbf{U}}_1, \overline{\mathbf{U}}_2, \overline{\mathbf{U}}_3\} = \operatorname{col}(\overline{\mathbf{U}}_1) + \operatorname{col}(\overline{\mathbf{U}}_2) + \operatorname{col}(\overline{\mathbf{U}}_3). \tag{B.1}$$

Note that for the above subspace, each column vector of $\widetilde{\mathbf{U}}$ is a basis vector. In addition, we define the sum of two subspaces $\mathbf{U}_1, \mathbf{U}_2$ as $\mathbf{U}_1 + \mathbf{U}_2 = \{\mathbf{u}_1 + \mathbf{u}_2 \mid \mathbf{u}_1 \in \mathbf{U}_1, \mathbf{u}_2 \in \mathbf{U}_2\}$. Obviously, $\widetilde{\mathbf{U}}$ is an orthonormal matrix with at most $3r$ columns.

Similarly, let $\overline{\mathbf{V}}_1, \overline{\mathbf{V}}_2, \overline{\mathbf{V}}_3$ be the right singular matrices of $\mathbf{X}, \mathbf{V}, \mathbf{H}_V$, respectively. Define $\widetilde{\mathbf{V}}$ as the matrix spanned by the column of $\overline{\mathbf{V}}_1, \overline{\mathbf{V}}_2$ and $\overline{\mathbf{V}}_3$ such that

$$\operatorname{col}(\widetilde{\mathbf{V}}) = \operatorname{span}\{\overline{\mathbf{V}}_1, \overline{\mathbf{V}}_2, \overline{\mathbf{V}}_3\} = \operatorname{col}(\overline{\mathbf{V}}_1) + \operatorname{col}(\overline{\mathbf{V}}_2) + \operatorname{col}(\overline{\mathbf{V}}_3), \tag{B.2}$$

where $\widetilde{\mathbf{V}}$ has at most $3r$ columns.

### B.1  Proof of Lemma 5.1

In order to prove the local curvature condition, we need to make use of the following lemmas. In Lemma B.1, we denote $\widetilde{\mathbf{Z}} \in \mathbb{R}^{(d_1+d_2) \times r}$ as $\widetilde{\mathbf{Z}} = [\mathbf{U}; -\mathbf{V}]$. Recall $\mathbf{Z} = [\mathbf{U}; \mathbf{V}]$, then we have $\|\mathbf{U}^\top \mathbf{U} - \mathbf{V}^\top \mathbf{V}\|_F^2 = \|\widetilde{\mathbf{Z}}^\top \mathbf{Z}\|_F^2$, and $\nabla_\mathbf{Z}(\|\mathbf{U}^\top\mathbf{U} - \mathbf{V}^\top\mathbf{V}\|_F^2) = 4\widetilde{\mathbf{Z}}\widetilde{\mathbf{Z}}^\top \mathbf{Z}$. We refer Wang et al. (2016) to readers for a detailed proof of Lemma B.1. Lemma B.2, proved in Section C.1, is a variation of the regularity condition of the sample loss function $\mathcal{L}_N$ (Tu et al., 2015), which is essential to derive the linear convergence rate in our main theorem.

**Lemma B.1.** (Wang et al., 2016) Let $\mathbf{Z}, \mathbf{Z}^* \in \mathbb{R}^{(d_1+d_2) \times r}$. Denote the optimal rotation with respect to $\mathbf{Z}$ as $\mathbf{R} = \operatorname{argmin}_{\widetilde{\mathbf{R}} \in \mathbb{Q}_r} \|\mathbf{Z} - \mathbf{Z}^* \widetilde{\mathbf{R}}\|_F$, and $\mathbf{H} = \mathbf{Z} - \mathbf{Z}^* \mathbf{R}$. Consider the gradient of the regularization term $\|\widetilde{\mathbf{Z}}^\top \mathbf{Z}\|_F^2$, we have

$$\langle \widetilde{\mathbf{Z}} \widetilde{\mathbf{Z}}^\top \mathbf{Z}, \mathbf{H} \rangle \geq \frac{1}{2} \|\widetilde{\mathbf{Z}}^\top \mathbf{Z}\|_F^2 - \frac{1}{2} \|\widetilde{\mathbf{Z}}^\top \mathbf{Z}\|_F \cdot \|\mathbf{H}\|_F^2,$$

where $\widetilde{\mathbf{Z}} = [\mathbf{U}; -\mathbf{V}]$.

**Lemma B.2.** Suppose the sample loss function $\mathcal{L}_N$ satisfies Conditions 4.3 and 4.4. For any rank-$r$ matrices $\mathbf{X}, \mathbf{Y} \in \mathbb{R}^{d_1 \times d_2}$, let the singular value decomposition of $\mathbf{X}$ be $\overline{\mathbf{U}}_1 \mathbf{\Sigma}_1 \overline{\mathbf{V}}_1^\top$, then we have

$$\langle \nabla \mathcal{L}_N(\mathbf{X}) - \nabla \mathcal{L}_N(\mathbf{Y}), \mathbf{X} - \mathbf{Y} \rangle \geq \frac{1}{4L} \|\widetilde{\mathbf{U}}^\top (\nabla \mathcal{L}_N(\mathbf{X}) - \nabla \mathcal{L}_N(\mathbf{Y}))\|_F^2$$
$$+ \frac{1}{4L} \|(\nabla \mathcal{L}_N(\mathbf{X}) - \nabla \mathcal{L}_N(\mathbf{Y})) \widetilde{\mathbf{V}}\|_F^2 + \frac{\mu}{2} \|\mathbf{X} - \mathbf{Y}\|_F^2,$$

where $\widetilde{\mathbf{U}} \in \mathbb{R}^{d_1 \times r_1}$ is an orthonormal matrix with $r_1 \leq 3r$ satisfying $\operatorname{col}(\overline{\mathbf{U}}_1) \subseteq \operatorname{col}(\widetilde{\mathbf{U}})$, and $\widetilde{\mathbf{V}} \in \mathbb{R}^{d_2 \times r_2}$ is an orthonormal matrix with $r_2 \leq 3r$ satisfying $\operatorname{col}(\overline{\mathbf{V}}_1) \subseteq \operatorname{col}(\widetilde{\mathbf{V}})$.

Now, we are ready to prove Lemma 5.1.

*Proof of Lemma 5.1.* According to (5.2), we have

$$\langle \nabla \widetilde{F}_N(\mathbf{Z}), \mathbf{H} \rangle = \underbrace{\langle \nabla_{\mathbf{U}} \mathcal{L}_N(\mathbf{U}\mathbf{V}^\top), \mathbf{H}_U \rangle + \langle \nabla_{\mathbf{V}} \mathcal{L}_N(\mathbf{U}\mathbf{V}^\top), \mathbf{H}_V \rangle}_{I_1} + \frac{1}{2} \underbrace{\langle \widetilde{\mathbf{Z}} \widetilde{\mathbf{Z}}^\top \mathbf{Z}, \mathbf{H} \rangle}_{I_2}, \quad \text{(B.3)}$$

where $\widetilde{\mathbf{Z}} = [\mathbf{U}; -\mathbf{V}]$. Recall that $\mathbf{X}^* = \mathbf{U}^* \mathbf{V}^{*\top}$, and $\mathbf{X} = \mathbf{U}\mathbf{V}^\top$. Note that $\nabla_{\mathbf{U}} \mathcal{L}_N(\mathbf{U}\mathbf{V}^\top) = \nabla \mathcal{L}_N(\mathbf{X}) \mathbf{V}$, and $\nabla_{\mathbf{V}} \mathcal{L}_N(\mathbf{U}\mathbf{V}^\top) = \nabla \mathcal{L}_N(\mathbf{X})^\top \mathbf{U}$. Thus, for the term $I_1$ in (B.3), we have

$$I_1 = \langle \nabla \mathcal{L}_N(\mathbf{X}), \mathbf{U}\mathbf{V}^\top - \mathbf{U}^* \mathbf{V}^{*\top} + \mathbf{H}_U \mathbf{H}_V^\top \rangle$$
$$= \underbrace{\langle \nabla \mathcal{L}_N(\mathbf{X}) - \nabla \mathcal{L}_N(\mathbf{X}^*), \mathbf{X} - \mathbf{X}^* + \mathbf{H}_U \mathbf{H}_V^\top \rangle}_{I_{11}} + \underbrace{\langle \nabla \mathcal{L}_N(\mathbf{X}^*), \mathbf{X} - \mathbf{X}^* + \mathbf{H}_U \mathbf{H}_V^\top \rangle}_{I_{12}}. \quad \text{(B.4)}$$

First, we consider the term $I_{11}$ in (B.4). Recall the definition of $\widetilde{\mathbf{U}}$ and $\widetilde{\mathbf{V}}$ in (B.1) and (B.2), respectively. According to Lemma B.2, we have

$$\langle \nabla \mathcal{L}_N(\mathbf{X}) - \mathcal{L}_N(\mathbf{X}^*), \mathbf{X} - \mathbf{X}^* \rangle \geq \frac{1}{4L} \|\widetilde{\mathbf{U}}^\top (\nabla \mathcal{L}_N(\mathbf{X}) - \nabla \mathcal{L}_N(\mathbf{X}^*))\|_F^2$$
$$+ \frac{1}{4L} \|(\nabla \mathcal{L}_N(\mathbf{X}) - \nabla \mathcal{L}_N(\mathbf{X}^*)) \widetilde{\mathbf{V}}\|_F^2 + \frac{\mu}{2} \|\mathbf{X} - \mathbf{X}^*\|_F^2. \quad \text{(B.5)}$$

Second, for the remaining term in $I_{11}$, we have

$$\left| \langle \nabla \mathcal{L}_N(\mathbf{X}) - \mathcal{L}_N(\mathbf{X}^*), \mathbf{H}_U \mathbf{H}_V^\top \rangle \right| = \left| \langle \widetilde{\mathbf{U}}^\top (\nabla \mathcal{L}_N(\mathbf{X}) - \nabla \mathcal{L}_N(\mathbf{X}^*)), \widetilde{\mathbf{U}}^\top \mathbf{H}_U \mathbf{H}_V^\top \rangle \right|$$
$$\leq \|\widetilde{\mathbf{U}}^\top (\nabla \mathcal{L}_N(\mathbf{X}) - \nabla \mathcal{L}_N(\mathbf{X}^*))\|_F \cdot \|\widetilde{\mathbf{U}}^\top\|_2 \cdot \|\mathbf{H}_U \mathbf{H}_V^\top\|_F$$
$$\leq \frac{1}{2} \|\widetilde{\mathbf{U}}^\top (\nabla \mathcal{L}_N(\mathbf{X}) - \nabla \mathcal{L}_N(\mathbf{X}^*))\|_F \cdot \|\mathbf{H}\|_F^2, \quad \text{(B.6)}$$

where the equality holds because $\widetilde{\mathbf{U}}\widetilde{\mathbf{U}}^\top\mathbf{H}_U = \mathbf{H}_U$, the first inequality holds because $|\langle\mathbf{A},\mathbf{B}\rangle| \leq \|\mathbf{A}\|_F \cdot \|\mathbf{B}\|_F$ and $\|\mathbf{A}\mathbf{B}\|_F \leq \|\mathbf{A}\|_2 \cdot \|\mathbf{B}\|_F$, and the second inequality holds because $2\|\mathbf{A}\mathbf{B}\|_F \leq \|\mathbf{A}\|_F^2 + \|\mathbf{B}\|_F^2$ and $\widetilde{\mathbf{U}}$ is orthonormal. Similarly, we have

$$\left|\langle\nabla\mathcal{L}_N(\mathbf{X}) - \mathcal{L}_N(\mathbf{X}^*), \mathbf{H}_U\mathbf{H}_V^\top\rangle\right| \leq \frac{1}{2}\|(\nabla\mathcal{L}_N(\mathbf{X}) - \nabla\mathcal{L}_N(\mathbf{X}^*))\widetilde{\mathbf{V}}\|_F \cdot \|\mathbf{H}\|_F^2. \tag{B.7}$$

Thus combining (B.6) and (B.7), we have

$$\left|\langle\nabla\mathcal{L}_N(\mathbf{X}) - \mathcal{L}_N(\mathbf{X}^*), \mathbf{H}_U\mathbf{H}_V^\top\rangle\right| \leq \frac{1}{4}\|\widetilde{\mathbf{U}}^\top(\nabla\mathcal{L}_N(\mathbf{X}) - \nabla\mathcal{L}_N(\mathbf{X}^*))\|_F \cdot \|\mathbf{H}\|_F^2$$
$$+ \frac{1}{4}\|(\nabla\mathcal{L}_N(\mathbf{X}) - \nabla\mathcal{L}_N(\mathbf{X}^*))\widetilde{\mathbf{V}}\|_F \cdot \|\mathbf{H}\|_F^2. \tag{B.8}$$

Therefore, combining (B.5) and (B.8), the term $I_{11}$ can be lower bounded by

$$I_{11} \geq \frac{1}{4L}\big(\|\widetilde{\mathbf{U}}^\top(\nabla\mathcal{L}_N(\mathbf{X}) - \nabla\mathcal{L}_N(\mathbf{X}^*))\|_F^2 + \|(\nabla\mathcal{L}_N(\mathbf{X}) - \nabla\mathcal{L}_N(\mathbf{X}^*))\widetilde{\mathbf{V}}\|_F^2\big) + \frac{\mu}{2}\|\mathbf{X} - \mathbf{X}^*\|_F^2$$
$$- \frac{1}{4}\big(\|\widetilde{\mathbf{U}}^\top(\nabla\mathcal{L}_N(\mathbf{X}) - \nabla\mathcal{L}_N(\mathbf{X}^*))\|_F + \|(\nabla\mathcal{L}_N(\mathbf{X}) - \nabla\mathcal{L}_N(\mathbf{X}^*))\widetilde{\mathbf{V}}\|_F\big) \cdot \|\mathbf{H}\|_F^2$$
$$\geq \frac{\mu}{2}\|\mathbf{X} - \mathbf{X}^*\|_F^2 - \frac{L}{8}\|\mathbf{H}\|_F^4, \tag{B.9}$$

where the last inequality holds because $2ab \leq ca^2 + b^2/c$, for any $c > 0$. Next, for the term $I_{12}$ in (B.4), we have

$$\left|\langle\nabla\mathcal{L}_N(\mathbf{X}^*), \mathbf{X} - \mathbf{X}^*\rangle\right| \leq \|\nabla\mathcal{L}_N(\mathbf{X}^*)\|_2 \cdot \|\mathbf{X} - \mathbf{X}^*\|_* \leq \sqrt{2r}\|\nabla\mathcal{L}_N(\mathbf{X}^*)\|_2 \cdot \|\mathbf{X} - \mathbf{X}^*\|_F, \tag{B.10}$$

where the first inequality is due to the Von Neumann trace inequality, and the second inequality is due to the fact that $\text{rank}(\mathbf{X} - \mathbf{X}^*) \leq 2r$. Similar for the remaining term in $I_{12}$, we have

$$\left|\langle\nabla\mathcal{L}_N(\mathbf{X}^*), \mathbf{H}_U\mathbf{H}_V^\top\rangle\right| \leq \sqrt{2r}\|\nabla\mathcal{L}_N(\mathbf{X}^*)\|_2 \cdot \|\mathbf{H}_U\mathbf{H}_V^\top\|_F. \tag{B.11}$$

Thus, combining (B.10) and (B.11), the term $I_{12}$ can be lower bounded by

$$I_{12} \geq -\sqrt{2r}\|\nabla\mathcal{L}_N(\mathbf{X}^*)\|_2 \cdot \left(\|\mathbf{X} - \mathbf{X}^*\|_F + \frac{1}{2}\|\mathbf{H}\|_F^2\right)$$
$$\geq -\frac{\mu}{8}\|\mathbf{X} - \mathbf{X}^*\|_F^2 - \frac{L}{4}\|\mathbf{H}\|_F^4 - \left(\frac{4r}{\mu} + \frac{r}{2L}\right) \cdot \|\nabla\mathcal{L}_N(\mathbf{X}^*)\|_2^2, \tag{B.12}$$

where the first inequality follows from the fact that $2\|\mathbf{A}\mathbf{B}\|_F \leq \|\mathbf{A}\|_F^2 + \|\mathbf{B}\|_F^2$, and the last inequality is due to $2ab \leq ca^2 + b^2/c$, for any $c > 0$. Therefore, plugging (B.9) and (B.12) into (B.4), we obtain the lower bound of $I_1$

$$I_1 \geq \frac{3\mu}{8}\|\mathbf{X} - \mathbf{X}^*\|_F^2 - \frac{3L}{8}\|\mathbf{H}\|_F^4 - \left(\frac{4r}{\mu} + \frac{r}{2L}\right) \cdot \|\nabla\mathcal{L}_N(\mathbf{X}^*)\|_2^2. \tag{B.13}$$

On the other hand, for the term $I_2$ in (B.3), according to lemma B.1, we have

$$I_2 \geq \frac{1}{2}\|\widetilde{\mathbf{Z}}^\top\mathbf{Z}\|_F^2 - \frac{1}{2}\|\widetilde{\mathbf{Z}}^\top\mathbf{Z}\|_F \cdot \|\mathbf{H}\|_F^2 \geq \frac{1}{4}\|\widetilde{\mathbf{Z}}^\top\mathbf{Z}\|_F^2 - \frac{1}{4}\|\mathbf{H}\|_F^4, \tag{B.14}$$

where the last inequality holds because $2ab \leq a^2 + b^2$. By plugging (B.13) and (B.14) into (B.3), we have

$$\langle \nabla \widetilde{F}_n(\mathbf{Z}), \mathbf{H} \rangle \geq \frac{3\mu}{8}\|\mathbf{X}-\mathbf{X}^*\|_F^2 + \frac{1}{8}\|\widetilde{\mathbf{Z}}^\top \mathbf{Z}\|_F^2 - \frac{3L+1}{8}\|\mathbf{H}\|_F^4 - \left(\frac{4r}{\mu} + \frac{r}{2L}\right) \cdot \|\nabla \mathcal{L}_N(\mathbf{X}^*)\|_2^2. \quad \text{(B.15)}$$

Furthermore, denote $\widetilde{\mathbf{Z}}^* = [\mathbf{U}^*; -\mathbf{V}^*]$, then we obtain

$$\begin{aligned}
\|\widetilde{\mathbf{Z}}^\top \mathbf{Z}\|_F^2 &= \langle \mathbf{ZZ}^\top - \mathbf{Z}^*\mathbf{Z}^{*\top}, \widetilde{\mathbf{Z}}\widetilde{\mathbf{Z}}^\top - \widetilde{\mathbf{Z}}^*\widetilde{\mathbf{Z}}^{*\top} \rangle + \langle \mathbf{Z}^*\mathbf{Z}^{*\top}, \widetilde{\mathbf{Z}}\widetilde{\mathbf{Z}}^\top \rangle + \langle \mathbf{ZZ}^\top, \widetilde{\mathbf{Z}}^*\widetilde{\mathbf{Z}}^{*\top} \rangle \\
&\geq \langle \mathbf{ZZ}^\top - \mathbf{Z}^*\mathbf{Z}^{*\top}, \widetilde{\mathbf{Z}}\widetilde{\mathbf{Z}}^\top - \widetilde{\mathbf{Z}}^*\widetilde{\mathbf{Z}}^{*\top} \rangle \\
&= \|\mathbf{U}\mathbf{U}^\top - \mathbf{U}^*\mathbf{U}^{*\top}\|_F^2 + \|\mathbf{V}\mathbf{V}^\top - \mathbf{V}^*\mathbf{V}^{*\top}\|_F^2 - 2\|\mathbf{U}\mathbf{V}^\top - \mathbf{U}^*\mathbf{V}^{*\top}\|_F^2, \quad \text{(B.16)}
\end{aligned}$$

where the first equality is due to $\widetilde{\mathbf{Z}}^{*\top}\mathbf{Z}^* = 0$, and the inequality is due to $\langle \mathbf{A}\mathbf{A}^\top, \mathbf{B}\mathbf{B}^\top \rangle = \|\mathbf{A}^\top \mathbf{B}\|_F^2 \geq 0$. Thus, according to Lemma D.2, we have

$$4\|\mathbf{X}-\mathbf{X}^*\|_F^2 + \|\widetilde{\mathbf{Z}}^\top \mathbf{Z}\|_F^2 = \|\mathbf{ZZ}^\top - \mathbf{Z}^*\mathbf{Z}^{*\top}\|_F^2 \geq 4(\sqrt{2}-1)\sigma_r \|\mathbf{H}\|_F^2, \quad \text{(B.17)}$$

where the first inequality holds because of (B.16), and the second inequality is due to Lemma D.2 and the fact that $\sigma_r^2(\mathbf{Z}^*) = 2\sigma_r$. Denote $\mu' = \min\{\mu, 1\}$. Therefore, plugging (B.17) into (B.15), we have

$$\begin{aligned}
\langle \nabla \widetilde{F}_N(\mathbf{Z}), \mathbf{H} \rangle \geq{}& \frac{\mu}{8}\|\mathbf{X}-\mathbf{X}^*\|_F^2 + \frac{\mu'\sigma_r}{10}\|\mathbf{H}\|_F^2 + \frac{1}{16}\|\widetilde{\mathbf{Z}}^\top \mathbf{Z}\|_F^2 \\
& - \frac{3L+1}{8}\|\mathbf{H}\|_F^4 - \left(\frac{4r}{\mu} + \frac{r}{2L}\right) \cdot \|\nabla \mathcal{L}_N(\mathbf{X}^*)\|_2^2,
\end{aligned}$$

which completes the proof. $\square$

## B.2 Proof of Lemma 5.2

*Proof.* Consider the term $\mathbf{G}_U$ first. Denote $\mathbf{X} = \mathbf{U}\mathbf{V}^\top$. According to the definition of $\mathbf{G}_U$, we have

$$\begin{aligned}
\|\mathbf{G}_U\|_F^2 &= \|\nabla_\mathbf{U} F_i(\mathbf{U}, \mathbf{V}) - \nabla \mathcal{L}_i(\widetilde{\mathbf{X}})\mathbf{V} + \nabla \mathcal{L}_N(\widetilde{\mathbf{X}})\mathbf{V}\|_F^2 \\
&= \left\|\nabla \mathcal{L}_i(\mathbf{X})\mathbf{V} - \nabla \mathcal{L}_i(\widetilde{\mathbf{X}})\mathbf{V} + \nabla \mathcal{L}_N(\widetilde{\mathbf{X}})\mathbf{V} + \frac{1}{2}\mathbf{U}(\mathbf{U}^\top \mathbf{U} - \mathbf{V}^\top \mathbf{V})\right\|_F^2 \\
&\leq 2\underbrace{\|\nabla \mathcal{L}_i(\mathbf{X})\mathbf{V} - \nabla \mathcal{L}_i(\widetilde{\mathbf{X}})\mathbf{V} + \nabla \mathcal{L}_N(\widetilde{\mathbf{X}})\mathbf{V}\|_F^2}_{I_1} + \frac{1}{2}\|\mathbf{U}^\top \mathbf{U} - \mathbf{V}^\top \mathbf{V}\|_F^2 \cdot \|\mathbf{U}\|_2^2, \quad \text{(B.18)}
\end{aligned}$$

where the second equality follows from definition of $\mathcal{F}_i$ in (3.5), and the inequality holds because $\|\mathbf{A}+\mathbf{B}\|_F^2 \leq 2\|\mathbf{A}\|_F^2 + 2\|\mathbf{B}\|_F^2$ and $\|\mathbf{AB}\|_F \leq \|\mathbf{A}\|_2 \cdot \|\mathbf{B}\|_F$. As for the term $I_1$ in (B.18), we further have

$$\begin{aligned}
I_1 &= \|\nabla \mathcal{L}_i(\mathbf{X})\mathbf{V} - \nabla \mathcal{L}_i(\widetilde{\mathbf{X}})\mathbf{V} + \nabla \mathcal{L}_N(\widetilde{\mathbf{X}})\mathbf{V} - \nabla \mathcal{L}_N(\mathbf{X})\mathbf{V} + \nabla \mathcal{L}_N(\mathbf{X})\mathbf{V} - \nabla \mathcal{L}_N(\mathbf{X}^*)\mathbf{V} + \nabla \mathcal{L}_N(\mathbf{X}^*)\mathbf{V}\|_F^2 \\
&\leq 3\|\widetilde{\mathbf{G}}_U\|_F^2 + 3\|\nabla \mathcal{L}_N(\mathbf{X})\mathbf{V} - \nabla \mathcal{L}_N(\mathbf{X}^*)\mathbf{V}\|_F^2 + 3\|\nabla \mathcal{L}_N(\mathbf{X}^*)\mathbf{V}\|_F^2 \\
&\leq 3\|\widetilde{\mathbf{G}}_U\|_F^2 + 3\|\nabla \mathcal{L}_N(\mathbf{X})\mathbf{V} - \nabla \mathcal{L}_N(\mathbf{X}^*)\mathbf{V}\|_F^2 + 3r\|\nabla \mathcal{L}_N(\mathbf{X}^*)\|_2^2 \cdot \|\mathbf{V}\|_2^2, \quad \text{(B.19)}
\end{aligned}$$

where we define $\widetilde{\mathbf{G}}_U = \nabla\mathcal{L}_i(\mathbf{X})\mathbf{V} - \nabla\mathcal{L}_N(\mathbf{X})\mathbf{V} - \nabla\mathcal{L}_i(\widetilde{\mathbf{X}})\mathbf{V} + \nabla\mathcal{L}_N(\widetilde{\mathbf{X}})\mathbf{V}$, the second inequality holds because $\|\mathbf{A} + \mathbf{B} + \mathbf{C}\|_F^2 \leq 3\|\mathbf{A}\|_F^2 + 3\|\mathbf{B}\|_F^2 + 3\|\mathbf{C}\|_F^2$, and the last inequality holds because $\|\mathbf{AB}\|_F \leq \|\mathbf{A}\|_2 \cdot \|\mathbf{B}\|_F$ and $\mathbf{V}$ has rank $r$. Thus combining (B.18) and (B.19), we have

$$\mathbb{E}\|\mathbf{G}_U\|_F^2 \leq 6\underbrace{\mathbb{E}\|\widetilde{\mathbf{G}}_U\|_F^2}_{I_2} + 6\underbrace{\|\nabla\mathcal{L}_N(\mathbf{X})\mathbf{V} - \nabla\mathcal{L}_N(\mathbf{X}^*)\mathbf{V}\|_F^2}_{I_3}$$
$$+ \frac{1}{2}\|\mathbf{U}^\top\mathbf{U} - \mathbf{V}^\top\mathbf{V}\|_F^2 \cdot \|\mathbf{U}\|_2^2 + 6r\|\nabla\mathcal{L}_N(\mathbf{X}^*)\|_2^2 \cdot \|\mathbf{V}\|_2^2, \qquad \text{(B.20)}$$

where the expectation is taken with respect to $i$. Next, we are going to upper bound $I_2$ and $I_3$, respectively. First, let us consider $I_2$ in (B.20). Since $i$ is uniformly picked from $[n]$, we have $\mathbb{E}[\nabla\mathcal{L}_i(\mathbf{X})\mathbf{V}] = \nabla\mathcal{L}_N(\mathbf{X})\mathbf{V}$ and $\mathbb{E}[\nabla\mathcal{L}_i(\widetilde{\mathbf{X}})\mathbf{V}] = \nabla\mathcal{L}_N(\widetilde{\mathbf{X}})\mathbf{V}$. Recall the definition of $\widetilde{\mathbf{V}}$ in (B.2), we have

$$\mathbb{E}\|\widetilde{\mathbf{G}}_U\|_F^2 = \mathbb{E}\big\|[\nabla\mathcal{L}_i(\mathbf{X})\mathbf{V} - \nabla\mathcal{L}_i(\widetilde{\mathbf{X}})\mathbf{V}] - \mathbb{E}[\nabla\mathcal{L}_i(\mathbf{X})\mathbf{V} - \nabla\mathcal{L}_i(\widetilde{\mathbf{X}})\mathbf{V}]\big\|_F^2$$
$$\leq \mathbb{E}\big\|\nabla\mathcal{L}_i(\mathbf{X})\mathbf{V} - \nabla\mathcal{L}_i(\widetilde{\mathbf{X}})\mathbf{V}\big\|_F^2$$
$$\leq \mathbb{E}\|(\nabla\mathcal{L}_i(\mathbf{X}) - \nabla\mathcal{L}_i(\widetilde{\mathbf{X}}))\widetilde{\mathbf{V}}\|_F^2 \cdot \|\widetilde{\mathbf{V}}\mathbf{V}\|_2^2$$
$$\leq \frac{1}{n}\sum_{i=1}^n \big\|(\nabla\mathcal{L}_i(\mathbf{X}) - \nabla\mathcal{L}_i(\widetilde{\mathbf{X}}))\widetilde{\mathbf{V}}\big\|_F^2 \cdot \|\mathbf{V}\|_2^2, \qquad \text{(B.21)}$$

where the first inequality holds because $\mathbb{E}\|\boldsymbol{\xi} - \mathbb{E}\boldsymbol{\xi}\|_2^2 \leq \mathbb{E}\|\boldsymbol{\xi}\|_2^2$ for any random vector $\boldsymbol{\xi}$, the second inequality holds because $\widetilde{\mathbf{V}}\widetilde{\mathbf{V}}^\top\mathbf{V} = 0$ and $\|\mathbf{AB}\|_F \leq \|\mathbf{A}\|_2 \cdot \|\mathbf{B}\|_F$, and the last inequality holds because $\|\mathbf{AB}\|_2 \leq \|\mathbf{A}\|_2 \cdot \|\mathbf{B}\|_2$ and $\|\widetilde{\mathbf{V}}\|_2 = 1$. Similarly, as for the term $I_3$ in (B.20), we have

$$I_3 = \big\|\big(\nabla\mathcal{L}_N(\mathbf{X}) - \nabla\mathcal{L}_N(\mathbf{X}^*)\big)\widetilde{\mathbf{V}}\widetilde{\mathbf{V}}^\top\mathbf{V}\big\|_F^2 \leq \big\|\big(\nabla\mathcal{L}_N(\mathbf{X}) - \nabla\mathcal{L}_N(\mathbf{X}^*)\big)\widetilde{\mathbf{V}}\big\|_F^2 \cdot \|\mathbf{V}\|_2^2, \qquad \text{(B.22)}$$

where the equality holds because $\widetilde{\mathbf{V}}\widetilde{\mathbf{V}}^\top\mathbf{V} = \mathbf{V}$, and the inequality holds because $\widetilde{\mathbf{V}}$ is orthonormal. Plugging (B.21) and (B.22) into (B.20), we obtain

$$\mathbb{E}\|\mathbf{G}_U\|_F^2 \leq \frac{6}{n}\sum_{i=1}^n \big\|\big(\nabla\mathcal{L}_i(\mathbf{X}) - \nabla\mathcal{L}_i(\widetilde{\mathbf{X}})\big)\widetilde{\mathbf{V}}\big\|_F^2 \cdot \|\mathbf{V}\|_2^2 + 6\big\|\big(\nabla\mathcal{L}_N(\mathbf{X}) - \nabla\mathcal{L}_N(\mathbf{X}^*)\big)\widetilde{\mathbf{V}}\big\|_F^2 \cdot \|\mathbf{V}\|_2^2$$
$$+ \frac{1}{2}\|\mathbf{U}^\top\mathbf{U} - \mathbf{V}^\top\mathbf{V}\|_F^2 \cdot \|\mathbf{U}\|_2^2 + 6r\|\nabla\mathcal{L}_N(\mathbf{X}^*)\|_2^2 \cdot \|\mathbf{V}\|_2^2. \qquad \text{(B.23)}$$

As for $\mathbb{E}\|\mathbf{G}_V\|_F^2$, by the same techniques, we have

$$\mathbb{E}\|\mathbf{G}_V\|_F^2 \leq \frac{6}{n}\sum_{i=1}^n \big\|\widetilde{\mathbf{U}}^\top\big(\nabla\mathcal{L}_i(\mathbf{X}) - \nabla\mathcal{L}_i(\widetilde{\mathbf{X}})\big)\big\|_F^2 \cdot \|\mathbf{U}\|_2^2 + 6\big\|\widetilde{\mathbf{U}}^\top\big(\nabla\mathcal{L}_N(\mathbf{X}) - \nabla\mathcal{L}_N(\mathbf{X}^*)\big)\big\|_F^2 \cdot \|\mathbf{U}\|_2^2$$
$$+ \frac{1}{2}\|\mathbf{U}^\top\mathbf{U} - \mathbf{V}^\top\mathbf{V}\|_F^2 \cdot \|\mathbf{V}\|_2^2 + 6r\|\nabla\mathcal{L}_N(\mathbf{X}^*)\|_2^2 \cdot \|\mathbf{U}\|_2^2, \qquad \text{(B.24)}$$

where $\widetilde{\mathbf{U}}$ is defined in (B.1). Recall $\mathbf{Z} = [\mathbf{U}; \mathbf{V}]$ and note that $\max\{\|\mathbf{U}\|_2, \|\mathbf{V}\|_2\} \leq \|\mathbf{Z}\|_2$. Thus combining (B.23) and (B.24), we obtain the upper bound of $\mathbb{E}\|\mathbf{G}\|_F^2$

$$\mathbb{E}\|\mathbf{G}\|_F^2 \leq \frac{12}{n} \sum_{i=1}^n \big( \underbrace{\|(\nabla\mathcal{L}_i(\mathbf{X}) - \nabla\mathcal{L}_i(\widetilde{\mathbf{X}}))\widetilde{\mathbf{V}}\|_F^2 + \|\widetilde{\mathbf{U}}^\top(\nabla\mathcal{L}_i(\mathbf{X}) - \nabla\mathcal{L}_i(\widetilde{\mathbf{X}}))\|_F^2}_{I_4} \big) \cdot \|\mathbf{Z}\|_2^2$$
$$+ 12 \big( \underbrace{\|(\nabla\mathcal{L}_N(\mathbf{X}) - \nabla\mathcal{L}_N(\mathbf{X}^*))\widetilde{\mathbf{V}}\|_F^2 + \|\widetilde{\mathbf{U}}^\top(\nabla\mathcal{L}_N(\mathbf{X}) - \nabla\mathcal{L}_N(\mathbf{X}^*))\|_F^2}_{I_5} \big) \cdot \|\mathbf{Z}\|_2^2$$
$$+ \|\mathbf{U}^\top\mathbf{U} - \mathbf{V}^\top\mathbf{V}\|_F^2 \cdot \|\mathbf{Z}\|_2^2 + 12r\|\nabla\mathcal{L}_N(\mathbf{X}^*)\|_2^2 \cdot \|\mathbf{Z}\|_2^2. \tag{B.25}$$

Finally, according to the Lemma C.1 and the restricted strong smoothness Conditions 4.4 and 4.6, we obtain the upper bound of $I_4$ and $I_5$

$$I_4 \leq 4L'\big(\mathcal{L}_i(\widetilde{\mathbf{X}}) - \mathcal{L}_i(\mathbf{X}) - \langle\nabla\mathcal{L}_i(\mathbf{X}), \widetilde{\mathbf{X}} - \mathbf{X}\rangle\big) \leq 2L'^2\|\widetilde{\mathbf{X}} - \mathbf{X}\|_F^2 \leq 4L'^2(\|\widetilde{\mathbf{X}} - \mathbf{X}^*\|_F^2 + \|\mathbf{X} - \mathbf{X}^*\|_F^2), \tag{B.26}$$

where the last inequality holds because $\|\mathbf{A} + \mathbf{B}\|_F^2 \leq 2\|\mathbf{A}\|_F^2 + 2\|\mathbf{B}\|_F^2$. Similarly we have

$$I_5 \leq 4L\big(\mathcal{L}_N(\mathbf{X}^*) - \mathcal{L}_N(\mathbf{X}) - \langle\nabla\mathcal{L}_N(\mathbf{X}), \mathbf{X}^* - \mathbf{X}\rangle\big) \leq 2L^2\|\mathbf{X}^* - \mathbf{X}\|_F^2. \tag{B.27}$$

Hence, plugging (B.26) and (B.27) into (B.25), we obtain

$$\mathbb{E}\|\mathbf{G}\|_F^2 \leq \big(48L'^2\|\widetilde{\mathbf{X}} - \mathbf{X}^*\|_F^2 + 24(2L'^2 + L^2)\|\mathbf{X} - \mathbf{X}^*\|_F^2\big) \cdot \|\mathbf{Z}\|_2^2$$
$$+ \big(\|\mathbf{U}^\top\mathbf{U} - \mathbf{V}^\top\mathbf{V}\|_F^2 + 12r\|\nabla\mathcal{L}_N(\mathbf{X}^*)\|_2^2\big) \cdot \|\mathbf{Z}\|_2^2,$$

which completes the proof. $\square$

## C  Proof of Technical Lemma in Appendix B

### C.1  Proof of Lemma B.2

In order to prove Lemma B.2, we need to make use of the following lemma. Lemma C.1 characterizes a projected version of restricted Lipschitz continuous gradient property regarding the sample loss function $\mathcal{L}_N$. In the following discussions, we first provide the proof of Lemma C.1.

**Lemma C.1.** Suppose the sample loss function $\mathcal{L}_N$ satisfies Conditions 4.3 and 4.4. For any rank-$r$ matrices $\mathbf{X}, \mathbf{Y} \in \mathbb{R}^{d_1 \times d_2}$, let the singular value decomposition of $\mathbf{X}$ be $\overline{\mathbf{U}}_1 \boldsymbol{\Sigma}_1 \overline{\mathbf{V}}_1^\top$, then we have

$$\mathcal{L}_N(\mathbf{Y}) \geq \mathcal{L}_N(\mathbf{X}) + \langle\nabla\mathcal{L}_N(\mathbf{X}), \mathbf{Y} - \mathbf{X}\rangle$$
$$+ \frac{1}{4L}\big(\|\widetilde{\mathbf{U}}^\top(\nabla\mathcal{L}_N(\mathbf{Y}) - \nabla\mathcal{L}_N(\mathbf{X}))\|_F^2 + \|(\nabla\mathcal{L}_N(\mathbf{Y}) - \nabla\mathcal{L}_N(\mathbf{X}))\widetilde{\mathbf{V}}\|_F^2\big),$$

where $\widetilde{\mathbf{U}} \in \mathbb{R}^{d_1 \times r_1}$ is an orthonormal matrix with $r_1 \leq 3r$ satisfying $\mathrm{col}(\overline{\mathbf{U}}_1) \subseteq \mathrm{col}(\widetilde{\mathbf{U}})$, and $\widetilde{\mathbf{V}} \in \mathbb{R}^{d_2 \times r_2}$ is an orthonormal matrix with $r_2 \leq 3r$ satisfying $\mathrm{col}(\overline{\mathbf{V}}_1) \subseteq \mathrm{col}(\widetilde{\mathbf{V}})$.

*Proof of Lemma C.1.* Define reference function $f_\mathbf{Y} : \mathbb{R}^{d_1 \times d_2} \to \mathbb{R}$ such that

$$f_\mathbf{Y}(\widetilde{\mathbf{X}}) = \mathcal{L}_N(\widetilde{\mathbf{X}}) - \mathcal{L}_N(\mathbf{Y}) - \langle\nabla\mathcal{L}_N(\mathbf{Y}), \widetilde{\mathbf{X}} - \mathbf{Y}\rangle. \tag{C.1}$$

Since $\mathcal{L}_N$ satisfies restricted strong convexity Condition 4.3, we have $f_{\mathbf{Y}}(\widetilde{\mathbf{X}}) \geq 0$, for any matrix $\widetilde{\mathbf{X}}$ with rank at most $3r$. Obviously, $f_{\mathbf{Y}}(\mathbf{Y}) = 0$. Recall the SVD of $\mathbf{X}$ is $\mathbf{X} = \overline{\mathbf{U}}_1 \mathbf{\Sigma}_1 \overline{\mathbf{V}}_1^\top$. Since $\text{col}(\overline{\mathbf{U}}_1) \subseteq \text{col}(\widetilde{\mathbf{U}})$ and $\overline{\mathbf{U}}_1^\top \overline{\mathbf{U}}_1 = \mathbf{I}_{r_1}$, we have $\widetilde{\mathbf{U}}\widetilde{\mathbf{U}}^\top \mathbf{X} = \mathbf{X}$. Thus we have

$$0 = f_{\mathbf{Y}}(\mathbf{Y}) \leq \min_\eta f_{\mathbf{Y}}\big(\widetilde{\mathbf{U}}\widetilde{\mathbf{U}}^\top [\mathbf{X} - \eta \nabla f_{\mathbf{Y}}(\mathbf{X})]\big)$$

$$= \min_\eta f_{\mathbf{Y}}\big(\mathbf{X} - \eta \widetilde{\mathbf{U}}\widetilde{\mathbf{U}}^\top \nabla f_{\mathbf{Y}}(\mathbf{X})\big)$$

$$\leq \min_\eta \Big\{ \mathcal{L}_N\big(\mathbf{X} - \eta \widetilde{\mathbf{U}}\widetilde{\mathbf{U}}^\top \nabla f_{\mathbf{Y}}(\mathbf{X})\big) - \mathcal{L}_N(\mathbf{Y}) - \langle \nabla \mathcal{L}_N(\mathbf{Y}), \mathbf{X} - \eta \widetilde{\mathbf{U}}\widetilde{\mathbf{U}}^\top \nabla f_{\mathbf{Y}}(\mathbf{X}) - \mathbf{Y} \rangle \Big\}, \tag{C.2}$$

where the first inequality holds because $\text{rank}(\mathbf{AB}) \leq \min\{\text{rank}(\mathbf{A}), \text{rank}(\mathbf{B})\}$ and $\widetilde{\mathbf{U}}\widetilde{\mathbf{U}}^\top$ has rank at most $3r$, and the last inequality follows from (C.1). Since $\mathcal{L}_N$ satisfies restricted strong smoothness Condition 4.4, we have

$$\mathcal{L}_N\big(\mathbf{X} - \eta \widetilde{\mathbf{U}}\widetilde{\mathbf{U}}^\top \nabla f_{\mathbf{Y}}(\mathbf{X})\big) \leq \mathcal{L}_N(\mathbf{X}) + \langle \nabla \mathcal{L}_N(\mathbf{X}), -\eta \widetilde{\mathbf{U}}\widetilde{\mathbf{U}}^\top \nabla f_{\mathbf{Y}}(\mathbf{X}) \rangle + \frac{L}{2} \| \eta \widetilde{\mathbf{U}}\widetilde{\mathbf{U}}^\top \nabla f_{\mathbf{Y}}(\mathbf{X}) \|_F^2. \tag{C.3}$$

Thus, plugging (C.3) into (C.2), we have

$$f_{\mathbf{Y}}(\mathbf{Y}) \leq \min_\eta \Big\{ f_{\mathbf{Y}}(\mathbf{X}) - \eta \langle \nabla \mathcal{L}_N(\mathbf{X}) - \nabla \mathcal{L}_N(\mathbf{Y}), \widetilde{\mathbf{U}}\widetilde{\mathbf{U}}^\top \nabla f_{\mathbf{Y}}(\mathbf{X}) \rangle + \frac{\eta^2 L}{2} \| \widetilde{\mathbf{U}}\widetilde{\mathbf{U}}^\top \nabla f_{\mathbf{Y}}(\mathbf{X}) \|_F^2 \Big\}$$

$$= f_{\mathbf{Y}}(\mathbf{X}) + \min_\eta \Big\{ -\eta \| \widetilde{\mathbf{U}}^\top \nabla f_{\mathbf{Y}}(\mathbf{X}) \|_F^2 + \frac{\eta^2 L}{2} \| \widetilde{\mathbf{U}}^\top \nabla f_{\mathbf{Y}}(\mathbf{X}) \|_F^2 \Big\}$$

$$= f_{\mathbf{Y}}(\mathbf{X}) - \frac{1}{2L} \| \widetilde{\mathbf{U}}^\top \nabla f_{\mathbf{Y}}(\mathbf{X}) \|_F^2, \tag{C.4}$$

where the first equality follows from (C.1), the second inequality holds because $\widetilde{\mathbf{U}}^\top \widetilde{\mathbf{U}} = \mathbf{I}_{r_1}$ and $\nabla f_{\mathbf{Y}}(\mathbf{X}) = \nabla \mathcal{L}_N(\mathbf{X}) - \nabla \mathcal{L}_N(\mathbf{Y})$, and the last equality holds because the minimizer is $\eta = 1/L$. Thus, plugging the definition of $f_{\mathbf{Y}}$ into (C.4), we obtain

$$\mathcal{L}_N(\mathbf{X}) - \mathcal{L}_N(\mathbf{Y}) - \langle \nabla \mathcal{L}_N(\mathbf{Y}), \mathbf{X} - \mathbf{Y} \rangle - \frac{1}{2L} \| \widetilde{\mathbf{U}}^\top (\nabla \mathcal{L}_N(\mathbf{X}) - \nabla \mathcal{L}_N(\mathbf{Y})) \|_F^2 \geq 0. \tag{C.5}$$

Since $\widetilde{\mathbf{V}}$ is orthonormal matrix with at most $3r$ columns and $\mathbf{X}\widetilde{\mathbf{V}}\widetilde{\mathbf{V}}^\top = \mathbf{X}$, following the same techniques, we obtain

$$\mathcal{L}_N(\mathbf{X}) - \mathcal{L}_N(\mathbf{Y}) - \langle \nabla \mathcal{L}_N(\mathbf{Y}), \mathbf{X} - \mathbf{Y} \rangle - \frac{1}{2L} \| (\nabla \mathcal{L}_N(\mathbf{X}) - \nabla \mathcal{L}_N(\mathbf{Y})) \widetilde{\mathbf{V}} \|_F^2 \geq 0. \tag{C.6}$$

Therefore, combining (C.5) and (C.6), we complete the proof. $\square$

Now we are ready to prove Lemma B.2.

*Proof of Lemma B.2.* By the restricted strong convexity of $\mathcal{L}_N$ in Condition 4.3, we have

$$\mathcal{L}_N(\mathbf{Y}) \geq \mathcal{L}_N(\mathbf{X}) + \langle \nabla \mathcal{L}_N(\mathbf{X}), \mathbf{Y} - \mathbf{X} \rangle + \frac{\mu}{2} \| \mathbf{X} - \mathbf{Y} \|_F^2. \tag{C.7}$$

Besides, according to lemma C.1, we have

$$\mathcal{L}_N(\mathbf{X}) - \mathcal{L}_N(\mathbf{Y}) \geq \langle \nabla \mathcal{L}_N(\mathbf{Y}), \mathbf{X} - \mathbf{Y} \rangle + \frac{1}{4L} \|\widetilde{\mathbf{U}}^\top (\nabla \mathcal{L}_N(\mathbf{X}) - \nabla \mathcal{L}_N(\mathbf{Y}))\|_F^2$$
$$+ \frac{1}{4L} \|(\nabla \mathcal{L}_N(\mathbf{X}) - \nabla \mathcal{L}_N(\mathbf{Y})) \widetilde{\mathbf{V}}\|_F^2. \tag{C.8}$$

Therefore, combining (C.7) and (C.8), we have

$$\langle \nabla \mathcal{L}_N(\mathbf{X}) - \nabla \mathcal{L}_N(\mathbf{Y}), \mathbf{X} - \mathbf{Y} \rangle \geq \frac{1}{4L} \|\widetilde{\mathbf{U}}^\top (\nabla \mathcal{L}_N(\mathbf{X}) - \nabla \mathcal{L}_N(\mathbf{Y}))\|_F^2$$
$$+ \frac{1}{4L} \|(\nabla \mathcal{L}_N(\mathbf{X}) - \nabla \mathcal{L}_N(\mathbf{Y})) \widetilde{\mathbf{V}}\|_F^2 + \frac{\mu}{2} \|\mathbf{X} - \mathbf{Y}\|_F^2,$$

which completes the proof. $\square$

## D  Auxiliary lemmas

For the completeness of our proofs, we provide several auxiliary lemmas in this section, which are originally proved in Tu et al. (2015).

**Lemma D.1.** (Tu et al., 2015) Assume $\mathbf{X}, \mathbf{Y} \in \mathbb{R}^{d_1 \times d_2}$ are two rank-$r$ matrices. Suppose they have singular value decomposition $\mathbf{X} = \mathbf{U}_1 \mathbf{\Sigma}_1 \mathbf{V}_1^\top$ and $\mathbf{Y} = \mathbf{U}_2 \mathbf{\Sigma}_2 \mathbf{V}_2^\top$. Suppose $\|\mathbf{X} - \mathbf{Y}\|_2 \leq \sigma_r(\mathbf{X})/2$, then we have

$$d^2\bigg([\mathbf{U}_2; \mathbf{V}_2] \mathbf{\Sigma}_1^{1/2}, [\mathbf{U}_1; \mathbf{V}_1] \mathbf{\Sigma}_2^{1/2}\bigg) \leq \frac{2}{\sqrt{2}-1} \frac{\|\mathbf{Y} - \mathbf{X}\|_F^2}{\sigma_r(\mathbf{X})}.$$

**Lemma D.2.** (Tu et al., 2015) For any matrices $\mathbf{Z}, \mathbf{Z}' \in \mathbb{R}^{(d_1+d_2) \times r}$, we have the following inequality

$$d^2(\mathbf{Z}, \mathbf{Z}') \leq \frac{1}{2(\sqrt{2}-1) \sigma_r^2(\mathbf{Z}')} \|\mathbf{Z}\mathbf{Z}^\top - \mathbf{Z}'\mathbf{Z}'^\top\|_F^2.$$

**Lemma D.3.** (Tu et al., 2015) For any matrices $\mathbf{Z}, \mathbf{Z}' \in \mathbb{R}^{(d_1+d_2) \times r}$, which satisfy $d(\mathbf{Z}, \mathbf{Z}') \leq \|\mathbf{Z}'\|_2 / 4$, we have the following inequality

$$\|\mathbf{Z}\mathbf{Z}^\top - \mathbf{Z}'\mathbf{Z}'^\top\|_F \leq \frac{9}{4} \|\mathbf{Z}'\|_2 \cdot d(\mathbf{Z}, \mathbf{Z}').$$